\documentclass[10pt,twocolumn,letterpaper]{article}

\PassOptionsToPackage{table,dvipsnames}{xcolor}
\usepackage{iccv}              
\usepackage{algorithm}
\usepackage{algorithmic}
\usepackage{array}
\usepackage{amsmath}
\usepackage{amssymb} 
\usepackage{enumitem}
\usepackage{multirow}
\usepackage{booktabs} 
\definecolor{iccvblue}{rgb}{0.21,0.49,0.74}
\usepackage[pagebackref,breaklinks,colorlinks,allcolors=iccvblue]{hyperref}

\def\paperID{8107} 
\def\confName{ICCV}
\def\confYear{2025}

\title{FRET: Feature Redundancy Elimination for Test Time Adaptation}

\author{
Linjing You\textsuperscript{1,*} \quad Jiabao Lu\textsuperscript{1,*} \quad Xiayuan Huang\textsuperscript{2,\dag} \quad Xiangli Nie\textsuperscript{1} \\
\textsuperscript{1}Institute of Automation, Chinese Academy of Sciences, Beijing, China\\
\textsuperscript{2}College of Science, Beijing Forestry University, Beijing, China\\
\texttt{\{youlinjing2023, lujiabao2025, xiangli.nie\}@ia.ac.cn}\\ \texttt{huangxiayuan@bjfu.edu.cn}
}

\begin{document}
\maketitle
\begin{abstract}

Test-Time Adaptation (TTA) aims to enhance the generalization of deep learning models when faced with test data that exhibits distribution shifts from the training data. In this context, only a pre-trained model and unlabeled test data are available, making it particularly relevant for privacy-sensitive applications. In practice, we observe that feature redundancy in embeddings tends to increase as domain shifts intensify in TTA. However, existing TTA methods often overlook this redundancy, which can hinder the model’s adaptability to new data. To address this issue, we introduce \textbf{F}eature \textbf{R}edundancy \textbf{E}limination for \textbf{T}est-time Adaptation (\textbf{FRET}), a novel perspective for TTA. A straightforward approach (S-FRET) is to directly minimize the feature redundancy score as an optimization objective to improve adaptation. Despite its simplicity and effectiveness, S-FRET struggles with label shifts, limiting its robustness in real-world scenarios. To mitigate this limitation, we further propose Graph-based FRET (G-FRET), which integrates a Graph Convolutional Network (GCN) with contrastive learning. This design not only reduces feature redundancy but also enhances feature discriminability in both the representation and prediction layers. Extensive experiments across multiple model architectures, tasks, and datasets demonstrate the effectiveness of S-FRET and show that G-FRET achieves state-of-the-art performance. Further analysis reveals that G-FRET enables the model to extract non-redundant and highly discriminative features during inference, thereby facilitating more robust test-time adaptation. The code is available at \href{https://anonymous.4open.science/r/fret-21BD}{https://anonymous.4open.science/r/fret-21BD}.

\end{abstract}    
\section{Introduction}

\begin{figure}[t]
\centering
\includegraphics[width=\columnwidth]{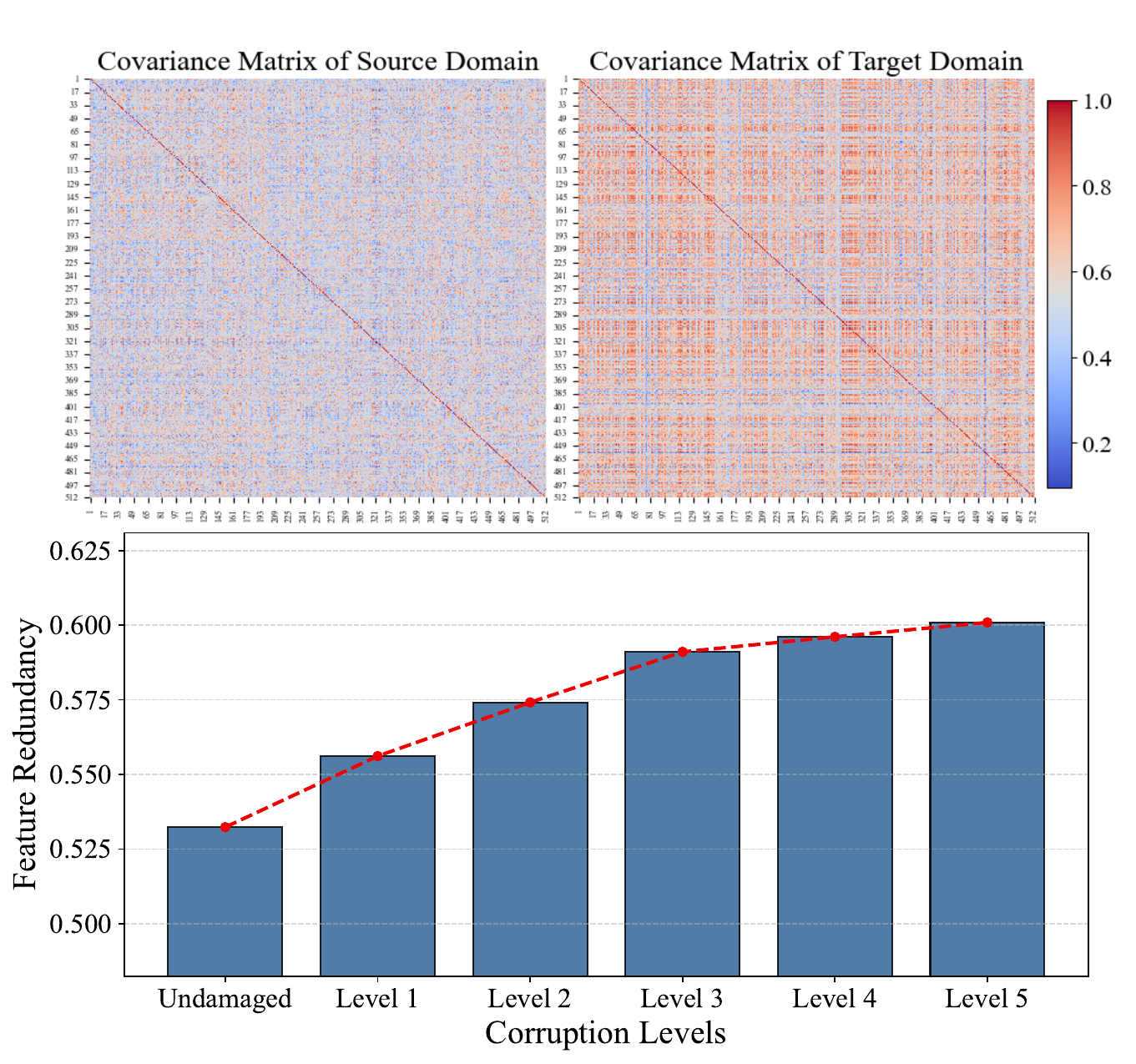} 
\caption{An intuitive demonstration of the relationship between embedded feature redundancy and distribution migration.\textbf{Top:} Second-order feature relation graph of embedding features. \textbf{Bottom:} Plot of feature redundancy versus the level of corruption.}
\label{fig:fig1}
\end{figure}

Recently, the emergence of deep neural networks (DNNs) \cite{deepLEARNING,he2016deep} has significantly propelled the advancement in a vast range of human tasks \cite{novo,alphago,alphafold}. However, the basic assumption that training and test data are drawn independently and identically (i.i.d.) may not be suitable in many real-world scenarios where the test/target distribution diverges from the training/source distribution \cite{wang2018deep}. This phenomenon, known as distribution shift, often leads to poor performance \cite{fang2020rethinking}. To tackle this issue, a variety of generalization or adaptation techniques have been developed to improve model robustness against distribution shifts \cite{you2019universal,zhou2022domain,liang2024comprehensive}. Among these, Test Time Adaptation (TTA) \cite{gong2024sotta,su2024towards,su2024revisiting} has been the focus of recent and active research. TTA has the advantage of only requiring access to the pre-trained model from the training/source domain and unlabeled test data. Therefore, TTA is a more secure and practical solution especially when models are publicly available but the training data and training process remain inaccessible due to privacy and resource restrictions \cite{liang2024comprehensive}.

Existing TTA methods can be broadly categorized into batch normalization calibration methods, pseudo-labeling methods, consistency training methods, and clustering-based training methods, based on different adaptation strategies \cite{liang2024comprehensive}. Batch normalization (BN) calibration methods \cite{wu2024test} posit that the statistics in the BN layers represent domain-specific knowledge. To bridge the domain gap, it is suggested to replace the training BN statistics with new statistics estimated over the entire target domain. Pseudo-labeling methods \cite{zeng2024rethinking} employ various filtering strategies, such as thresholding or entropy-based approaches, to obtain reliable pseudo-labels. This process helps minimize discrepancies between predicted labels and pseudo-labels. Consistency training methods \cite{Sinha_2023_WACV} aim to improve the stability of network predictions or features by addressing variations in input data, such as noise or perturbations, and changes in model parameters. Clustering-based training methods \cite{lee2024entropy} focus on reducing uncertainty in target network predictions by leveraging clustering techniques to organize target features into distinct groups, thereby enhancing the robustness of the model.

Despite the empirical effectiveness of these methods, none of them addresses the issue of feature redundancy in embeddings, which can impair the model's ability to generalize to new data \cite{li2017feature}. As depicted in \cref{fig:fig1}, the feature redundancy in ResNet-18's \cite{he2016deep} embeddings on the CIFAR10-C dataset \cite{hendrycks2019benchmarking} during distribution shifts is illustrated by using a second-order feature relations graph (i.e., covariance matrix of features) \cite{mh2024lvm}. Specifically, ResNet-18 transforms each image into a low-dimensional embedding with 512 features. $n$ embeddings are stacked into an $n \times 512$ matrix, and the absolute values of its covariance matrix ($512 \times 512$) are computed for visualization. In the resulting heatmap, deeper red colors indicate higher correlation between the corresponding features. Overall, a redder hue suggests higher redundancy within the feature set. The top two images in \cref{fig:fig1} shows the second-order feature relation graphs for the source and target domains. Visual inspection reveals that feature redundancy is significantly higher in the target domain compared to the source domain, which highlights the impact of distribution shift on feature redundancy. To further illustrate the relationship between feature redundancy and distribution shift, we remove the diagonal elements of the covariance matrix and calculate the sum of the absolute values of the remaining elements to derive the redundancy scores (see \cref{eq:L_{R_e}}). As depicted in the bottom of \cref{fig:fig1}, feature redundancy increases with the level of corruption (i.e., the degree of distribution shift). This suggests that \textbf{features with lower redundancy may be better suited for adaptation to the target domain}.

Building on the limitations identified in the existing literature and the observations above, we provide a novel perspective for test-time adaptation—\underline{F}eature \underline{R}edundancy \underline{E}limination for \underline{T}TA (\textbf{FRET})—which differs from existing methods. Due to the inherently differentiable nature of the redundancy metric, a straightforward and computationally efficient approach is to directly minimize the feature redundancy score. We refer to this simple approach as S-FRET. However, since S-FRET lacks the ability to perceive label distributions, it effectively addresses \textbf{covariate shift} but fails to handle \textbf{label shift}—a scenario where the target domain exhibits a label distribution that deviates from the source domain (e.g., long-tailed data). This discrepancy can significantly degrade model performance and remains a critical challenge~\cite{Park2023LabelSA, kang2019decoupling, cao2019learning}. 

To overcome S-FRET’s limitation in handling label shift, we further propose G-FRET, which integrates a \underline{G}raph Convolutional Network (GCN). Unlike FRET, which only minimizes redundancy at the representation layer, G-FRET leverages GCN to incorporate both attention and redundant feature relationships into the representation and prediction layers. Specifically, G-FRET learns non-redundant and discriminative representation through contrastive distillation between attention and redundant representation in the representation-layer. In the prediction layer, attention predictions are optimized through entropy minimization, while ensuring that they are distinct from redundant predictions by negative learning. By maximizing the discriminability of attention relationships while eliminating redundancy in both the representation and prediction layers, G-FRET effectively reduces test-time feature redundancy and improves discrimination when handling complex distribution-shifted test data. 

We summarize the contributions of this paper as follows.

\begin{itemize}
    \item \textbf{Potential Way:} provide a novel perspective for test-time adaptation — \underline{F}eature \underline{R}edundancy \underline{E}limination for \underline{T}TA (\textbf{FRET}) — which differs from existing methods. It adapts the model to distribution shifts by reducing the redundancy of embedded features during testing. To the best of our knowledge, this is the first attempt to leverage feature redundancy for TTA.
    
    \item \textbf{Innovative Methods:} We propose two novel methods, S-FRET and G-FRET, to eliminate feature redundancy in test-time embeddings. S-FRET directly minimizes the feature redundancy score, offering a lightweight solution. G-FRET extends this by decomposing feature relations into attention and redundancy components, enabling joint optimization of redundancy elimination and class discriminability in real-world scenarios.
    
    \item \textbf{Extensive Evaluation:} Comprehensive experiments across multiple model architectures, tasks, and datasets demonstrate the effectiveness of S-FRET and show that G-FRET achieves state-of-the-art performance. Further in-depth analysis reveals that G-FRET facilitates the extraction of non-redundant and discriminative embedding features during testing, ultimately enhancing model adaptation to the test-time target domain.
\end{itemize}
\section{Preliminary and Notations}
\label{sec:Related Work}

We briefly revisit Test-Time Adaptation and Feature Redundancy Elimination in this section for the convenience of introducing methods, and put detailed related work discussions into \cref{sec:Extended Related work} due to page limits.

\subsection{Test-Time Adaptation}

In the test-time adaptation (TTA) scenario, we have access only to unlabeled data from the target domain in an online manner and a pre-trained model from the source domain. Specifically, let $\{(x_i, y_i)\}_{i=1}^{n_s}\subset \mathcal{X}_s\times \mathcal{Y}_s$ represent the labeled source domain dataset with labels $\mathcal{Y}_s$ and $\{x_j\}_{j=1}^{n_t}\subset \mathcal{X}_t$ represent a batch of unlabeled target data. The model, trained on the source domain dataset and parameterized by $\theta$, is denoted as $g_\theta=h(f(\cdot)): \mathcal{X}_s\to\mathcal{Y}_s$, where $f$ is the backbone encoder and $h$ denotes the linear classification head. Our objective is to adapt this model to the target domain $\mathcal{X}_t$, which exhibits a different distribution from that of the source domain $\mathcal{X}_s$. During testing, for each instance $x_j\in\mathcal{X}_t$, let the output of $f$ and $h$ be denoted as embedding $z_j = f(x_j) \in \mathbb{R}^d$ and logits $p_j=h(z_j)\in \mathbb{R}^C$, respectively, where $d$ is the dimension of the embeddings and $C$ is the number of classes.

\subsection{Feature Redundancy Elimination}

Feature redundancy is a key concern in both feature extraction and feature selection \cite{FE_review2,FS_review3}.  Following previous work \cite{second_order}, we remove the diagonal elements of the covariance matrix and calculate the sum of the absolute values of the remaining elements to derive the redundancy scores $R_e$ as:
\begin{equation} \label{eq:L_{R_e}}
R_e = \left\lVert \tilde{Z}^T \tilde{Z} - I_d \right\rVert_1
\end{equation}
where $I_d$ is the identity matrix, $\lVert \cdot \rVert_1$ denotes the 1-norm of a matrix and $\tilde{Z}_{:,j} = \frac{Z_{:,j}}{\lVert Z_{:,j} \rVert_2}$ denotes the matrix obtained by normalizing each column.

\section{The Proposed Approach}

\begin{figure*}[t]
\centering
\includegraphics[width=\textwidth]{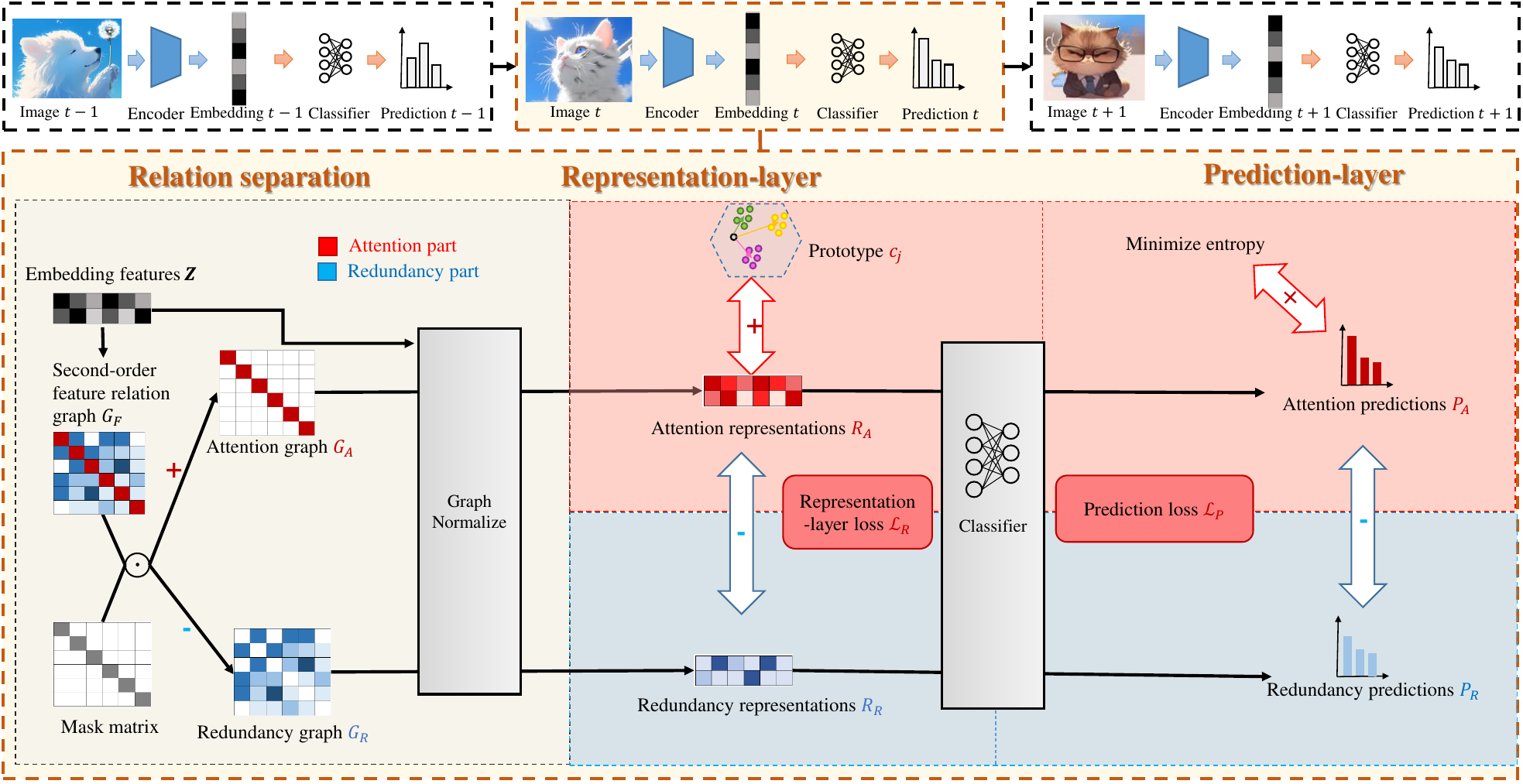} 
\caption{The pipeline of our proposed Graph Convolutional Network based Feature Redundancy Elimination for Test Time Adaptation (G-FRET) method. For each test sample, G-FRET learns non-redundant and discriminative representation through contrastive distillation in the representation-layer. Additionally, in the prediction layer, attention predictions are optimized through entropy minimization, while ensuring that they are distinct from redundant predictions by using negative learning.}
\label{fig:fig2}
\end{figure*}

\subsection{S-FRET}
To eliminate feature redundancy in the target domain, a simple yet effective approach is to directly minimize the feature redundancy $R_e$, as defined in  \cref{eq:L_{R_e}}, leading to the proposed method S-FRET (Simple and Straightforward FRET). The loss function for S-FRET can be written as:

\begin{equation}\label{eq:L_FRET}
\mathcal{L}_{SFRET} = \left\lVert \tilde{Z}^T \tilde{Z} - I_d \right\rVert_1
\end{equation}

This method directly minimizes the redundancy score derived from feature correlation, providing a computationally efficient solution that avoids complex architectural modifications. By focusing on the feature space of representation-layer, S-FRET enables the model to extract non-redundant representations, thereby enhancing its adaptability to the target domain (see \cref{fig:fig3}).

However, since the redundancy score does not incorporate information about class distributions, S-FRET primarily addresses \textbf{covariate shift} but lacks the discriminative capability to handle \textbf{label shift}—a critical challenge where the target domain exhibits a label distribution that deviates significantly from the source domain (e.g., long-tailed data), leading to substantial performance degradation (see \cref{tab:LS2}) \cite{Park2023LabelSA, kang2019decoupling, cao2019learning}.

\subsection{G-FRET}
\subsubsection{Overview of G-FRET}
To eliminate feature redundancy and address label shift in the target domain, we further propose a novel two-layer TTA method, namely \underline{G}raph Convolutional Network based \underline{F}eature \underline{R}edundancy \underline{E}limination for \underline{T}est Time Adaptation (G-FRET), which leverages GCN to simultaneously model feature relationships and sample-specific discriminative information in both representation and prediction layers. \cref{fig:fig2} provides an overview of our G-FRET framework.

The key idea of G-FRET is to adapt the pre-tarined model during testing by extracting discriminative and non-redundant embedding features from both the representation and prediction layers. Firstly (\cref{subsec: Attention Part and Redundancy Part}), we apply a mask matrix to the second-order feature relation graph. This process yields two separate graphs: an attention graph, which highlights significant feature relationships, and a redundant graph, which identifies redundant feature relationships. Next, we derive the attention/redundancy representation and attention/redundancy prediction by utilizing a single-layer GCN to integrate the model's embeddings with the attention and redundant graphs. This process allows us to explicitly separate attention and redundancy relations in both representation and prediction layer. Then (\cref{subsec: Representation-Layer Redundancy Elimination}), in the representation layer, contrastive learning techniques are employed to enhance the discriminability of the attention representation while ensuring it remains distant from the redundant representation. At last (\cref{subsec: Prediction-Layer Redundancy Elimination}), in the prediction layer, attention predictions are optimized through entropy minimization, while ensuring that they are distinct from redundant predictions by negative learning. Through this two-layer process, we achieve discriminative and non-redundant embedding features that effectively adapt to the target domain.

\subsubsection{Attention Part and Redundancy Part}
\label{subsec: Attention Part and Redundancy Part}
Given a batch of $n_t$ data instances from the target domain, we obtain the first-order $n_t \times d$ embedding feature matrix $Z \in \mathbb{R}^{n_t \times d}$ and the second-order feature relation graph $G_F = Z^T Z \in \mathbb{R}^{d \times d}$. Following the state-of-the-art feature selection method SOFT \cite{second_order}, we apply a symmetric mask matrix $M_M \in \mathbb{R}^{d \times d}$ to $G_F$.
This decomposition of $G_F$ yields two distinct graphs: the attention graph $G_A$ and the redundancy graph $G_R$. $G_A$ captures the attention relations (i.e., important relation) among features, while $G_R$ represents the redundant relation between them. Formally, they are calculated as follows:
\begin{equation}
G_A = G_F \odot M_M; G_R = G_F - G_A
\end{equation}
where $\odot$ is the element-wise product. However, differently from SOFT, which focuses on feature selection by deriving an optimal mask matrix $M_M$, our objective is to extract non-redundant features $Z$. For ideally non-redundant features, the second-order feature relation graph should be a diagonal matrix, indicating that each feature is independent of all others. Therefore, we set $M_M = I_d$ to ensure that $G_A$ captures only the essential feature relations, while $G_R$ includes all remaining relations as redundant relations. Here, $I_d \in \mathbb{R}^{d \times d}$ is the identity matrix. In addition, in some scenarios with prior knowledge, such as knowing which features may be more important, or needing to maintain special structural relations between features, some adaptations can be made by simply modifying the mask matrix $M_M$.

Then, in order to obtain attention/redundancy representation and attention/redundancy prediction based on the attention and redundancy graphs, we combine the normalized graph matrix with the classifier head $h$ to form a one-layer Graph Convolutional Network (GCN \cite{kipf2016semi}), which can help generate attention representations $R_A$ and attention predictions $P_A$ that contain both data information and feature relationship information:
\begin{equation}
R_A = Z D_A^{-1/2} G_A D_A^{-1/2}; P_A = R_A \theta^h
\end{equation}
where $D_A^{-1/2} G_A D_A^{-1/2}$ is the normalized graph matrix of graph $G_A$, $D_A$ is the degree matrix of the graph, and $\theta^h$ are the parameters of the classifier head $h$. Similarly, we can obtain redundancy representations $R_R$ and predictions $P_R$ from redundancy graph $G_R$:
\begin{equation}
R_R = Z D_R^{-1/2} G_R D_R^{-1/2};P_R = R_R \theta^h
\end{equation}

Therefore, we decompose the model's original presentation and prediction into two components: the attention part and the redundancy part. This decomposition enables a distinct focus on important relationships while identifying redundant ones. Specifically, $R_A$ and $P_A$ focus solely on the feature relationships within the attention part, while $R_R$ and $P_R$ are concerned exclusively with the feature relationships within the redundancy part. Based on this decomposition, our goal is to ensure that the attention part adapts effectively to the target domain. Simultaneously, we aim to separate redundancy part from the attention component in both the representation and prediction layers.

\subsubsection{Representation-Layer Redundancy Elimination}
\label{subsec: Representation-Layer Redundancy Elimination}
With the above representations $R_A$ and $R_R$, we expect that the attention representations possess class discriminability to handle label shift, while remaining distinct from redundancy representations to eliminate feature redundancy. To achieve this, we leverage contrastive learning \cite{PCL,INfo1} as follows:

\begin{align}\label{eq:L_R}
& \mathcal{L}_R =\notag \\
& -\sum_{i=1}^{n_t} \log \frac{\exp(\text{sim}(R_{A_i}, c_o))}{\sum_{j=1}^C \exp(\text{sim}(R_{A_i}, c_j)) + \exp(\text{sim}(R_{A_i}, R_{R_i}))}
\end{align}
where $\text{sim}(\cdot, \cdot)$ denotes cosine similarity, $\{c_j\}_{j=1}^C$ denote the class centers in the attention representation space, $C$ is the number of clusters, and $c_o$ represents the center of the cluster that contains $R_{A_i}$. $c_j$ is calculated as follows:
\begin{equation}
c_j = \frac{\sum_i z_i \cdot \mathbf{1}[\text{argmax}(p_i) = j]}{\sum_i \mathbf{1}[\text{argmax}(p_i) = j]}
\end{equation}
where $\mathbf{1}(\cdot)$ is an indicator function, outputting value 1 if the argument is true or 0 otherwise. 

By minimizing $\mathcal{L}_R$, the model simultaneously enhances the discriminability of attention representations while ensuring their distinction from redundancy representations. This process enables the model to extract robust, discriminative, and non-redundant embedding features in the representation layer, facilitating effective adaptation to the target domain.

\subsubsection{Prediction-Layer Redundancy Elimination}
\label{subsec: Prediction-Layer Redundancy Elimination}
In the prediction layer, to enhance the discriminability of attention predictions while reducing the influence of redundant predictions, we combine the principles of entropy minimization and negative learning into a unified loss function:
\begin{equation}\label{eq:L_P}
\mathcal{L}_P = -\sum_{i=1}^N \sigma(P_{A_i}) \log \sigma(P_{A_i})-\sum_{i=1}^N \sigma(P_{R_i}) \log \sigma(1 - P_{A_i})
\end{equation}
where $\sigma$ denotes the softmax operation. Here, the first term minimizes the entropy of attention predictions $P_A$, encouraging sharper and more confident predictions, while the second term applies negative learning to redundant predictions $P_R$, penalizing them to ensure they remain distinct from the attention predictions. 

\subsubsection{Training Loss Function}
By combining \cref{eq:L_R} and \cref{eq:L_P}, our overall loss function for G-FRET can be written as:

\begin{equation}\label{eq:L}
\mathcal{L}_{GFRET} = \mathcal{L}_R + \lambda \mathcal{L}_P
\end{equation}
where $\lambda$ is the balancing hyperparameter that controls the trade-off between $\mathcal{L}_R$ and $\mathcal{L}_P$.

During testing, adaptation is performed online: upon receiving instances $\{x_i\}_{i=1}^{n_t}$ at time $t$, the model is updated with parameters from the previous instances. The model then generates the prediction $\{p_i = g(x_i)\}_{i=1}^{n_t}$ for the new instances and updates using \cref{eq:L} with a single step of gradient descent. Adaptation continues as long as test data is available.
\begin{table*}[t!]
\begin{center}
\small
\resizebox{\textwidth}{!}{
\begin{tabular}{@{}l|l|cccc>{\columncolor{blue!10}}c|cccc>{\columncolor{blue!10}}l@{}}
\toprule
\multirow{2}{*}{Backbone} & \multirow{2}{*}{Method}   & \multicolumn{4}{c}{PACS}                                          & \multicolumn{1}{c}{\multirow{2}{*}{Avg}} & \multicolumn{4}{c}{OfficeHome}                                    & \multicolumn{1}{c}{\multirow{2}{*}{Avg}} \\ \cmidrule(lr){3-6} \cmidrule(lr){8-11}
                          &                           & A              & C              & P              & S              & \multicolumn{1}{c}{}           & A              & C              & P              & R              & \multicolumn{1}{c}{}                     \\ \midrule
\multirow{9}{*}{ResNet-18} & \multicolumn{1}{l|}{Source \cite{he2016deep}} &$78.37_{\pm 0.00}$ & $77.39_{\pm 0.00}$ & $95.03_{\pm 0.00}$ & $76.58_{\pm 0.00}$ & $81.84_{\pm 0.00}$ & $56.45_{\pm 0.00}$ & $48.02_{\pm 0.00}$ & $71.34_{\pm 0.00}$ & $72.23_{\pm 0.00}$ & $62.01_{\pm 0.00}$  \\
                          & \multicolumn{1}{l|}{BN \cite{BN}}     & $81.03_{\pm 0.09}$ & $80.67_{\pm 0.27}$ & $95.14_{\pm 0.15}$ & $73.83_{\pm 0.18}$ & $82.66_{\pm 0.08}$ & $55.66_{\pm 0.17}$ & $49.23_{\pm 0.32}$ & $70.73_{\pm 0.10}$ & $72.49_{\pm 0.19}$ & $62.03_{\pm 0.12}$  \\
                          & \multicolumn{1}{l|}{SAR \cite{SAR}}    & $83.62_{\pm 0.42}$ & $81.47_{\pm 0.93}$ & $95.14_{\pm 0.15}$ & $80.44_{\pm 0.74}$ & $85.17_{\pm 0.26}$ & $57.30_{\pm 0.44}$ & $\underline{50.54}_{\pm 0.26}$ & $70.71_{\pm 0.55}$ & $72.75_{\pm 0.27}$ & $62.82_{\pm 0.31}$  \\
                          & \multicolumn{1}{l|}{EATA \cite{EATA}}   & $82.06_{\pm 0.85}$ & $81.11_{\pm 1.03}$ & $95.16_{\pm 0.29}$ & $73.39_{\pm 1.00}$ & $82.93_{\pm 0.46}$ & $56.42_{\pm 0.41}$ & $49.75_{\pm 0.41}$ & $71.50_{\pm 0.35}$ & $72.85_{\pm 0.41}$ & $62.63_{\pm 0.11}$  \\
                          & \multicolumn{1}{l|}{TENT \cite{TENT}}   & $83.54_{\pm 1.78}$ & $82.26_{\pm 1.69}$ & $95.54_{\pm 0.31}$ & $81.05_{\pm 1.78}$ & $85.60_{\pm 0.75}$ & $57.25_{\pm 0.32}$ & $50.45_{\pm 0.26}$ & $72.18_{\pm 0.38}$ & $73.09_{\pm 0.34}$ & $63.24_{\pm 0.13}$  \\
                          & \multicolumn{1}{l|}{TSD \cite{TSD}}    & $\underline{87.05}_{\pm 0.47}$ & $\underline{86.94}_{\pm 0.62}$ & $95.98_{\pm 0.25}$ & $82.56_{\pm 1.12}$ & $\underline{88.13}_{\pm 0.44}$ & $\textbf{57.80}_{\pm 0.34}$ & $50.05_{\pm 0.69}$ & $71.45_{\pm 0.42}$ & $70.92_{\pm 0.46}$ & $62.55_{\pm 0.06}$  \\
                          & \multicolumn{1}{l|}{TEA \cite{TEA}}    & $87.04_{\pm 0.54}$ & $86.54_{\pm 0.85}$ & $\underline{96.17}_{\pm 0.36}$ & $82.17_{\pm 1.08}$ & $87.98_{\pm 0.48}$ & $57.07_{\pm 0.88}$ & $50.47_{\pm 0.18}$ & $71.84_{\pm 0.18}$ & $72.85_{\pm 0.10}$ & $63.06_{\pm 0.27}$  \\
                          & \multicolumn{1}{l|}{TIPI \cite{TIPI}}   & $84.79_{\pm 1.18}$ & $84.24_{\pm 1.01}$ & $95.86_{\pm 0.18}$ & $\textbf{84.05}_{\pm 0.75}$ & $87.23_{\pm 0.41}$ & $57.23_{\pm 0.34}$ & $50.42_{\pm 0.27}$ & $\underline{72.23}_{\pm 0.24}$ & $\underline{73.28}_{\pm 0.19}$ & $\underline{63.29}_{\pm 0.05}$  \\
                          \rowcolor[HTML]{FADADD}
                          & \multicolumn{1}{l|}{S-FRET}   & ${86.25}_{\pm 0.30}$ & ${86.66}_{\pm 0.37}$ & ${95.96}_{\pm 0.25}$ & ${75.90}_{\pm 2.44}$ & ${86.20}_{\pm 0.66}$ & ${56.26}_{\pm 0.21}$ & ${49.86}_{\pm 0.38}$ & ${71.65}_{\pm 0.05}$ & ${72.72}_{\pm 0.14}$ & ${62.62}_{\pm 0.09}$\\
                          \rowcolor[HTML]{FADADD}
                          & \multicolumn{1}{l|}{G-FRET}   & $\textbf{87.20}_{\pm 0.35}$ & $\textbf{87.19}_{\pm 0.24}$ & $\textbf{96.28}_{\pm 0.36}$ & $\underline{83.36}_{\pm 1.43}$ & $\textbf{88.51}_{\pm 0.39}$ & $\underline{57.49}_{\pm 0.34}$ & $\textbf{51.28}_{\pm 0.42}$ & $\textbf{73.04}_{\pm 0.29}$ & $\textbf{73.40}_{\pm 0.35}$ & $\textbf{63.81}_{\pm 0.11}$ \\ \midrule
\multirow{9}{*}{ResNet-50} & \multicolumn{1}{l|}{Source \cite{he2016deep}}& $83.89_{\pm 0.00}$ & $81.02_{\pm 0.00}$ & $96.17_{\pm 0.00}$ & $78.04_{\pm 0.00}$ & $84.78_{\pm 0.00}$ & $64.85_{\pm 0.00}$ & $52.26_{\pm 0.00}$ & $75.04_{\pm 0.00}$ & $75.88_{\pm 0.00}$ & $67.01_{\pm 0.00}$  \\
                          & \multicolumn{1}{l|}{BN \cite{BN}}     & $85.19_{\pm 0.35}$ & $85.49_{\pm 0.30}$ & $96.60_{\pm 0.19}$ & $72.25_{\pm 0.33}$ & $84.88_{\pm 0.08}$ & $63.30_{\pm 0.20}$ & $53.02_{\pm 0.32}$ & $73.59_{\pm 0.23}$ & $75.10_{\pm 0.29}$ & $66.25_{\pm 0.05}$  \\
                          & \multicolumn{1}{l|}{SAR \cite{SAR}}    & $85.24_{\pm 0.37}$ & $85.49_{\pm 0.30}$ & $96.60_{\pm 0.19}$ & $76.02_{\pm 0.84}$ & $85.84_{\pm 0.26}$ & $64.78_{\pm 0.17}$ & $56.30_{\pm 0.57}$ & $74.83_{\pm 0.27}$ & $76.18_{\pm 0.30}$ & $68.02_{\pm 0.19}$  \\
                          & \multicolumn{1}{l|}{EATA \cite{EATA}}   & $85.14_{\pm 0.31}$ & $85.36_{\pm 0.20}$ & $96.54_{\pm 0.19}$ & $72.27_{\pm 0.11}$ & $84.83_{\pm 0.11}$ & $64.25_{\pm 0.41}$ & $54.09_{\pm 0.53}$ & $74.25_{\pm 0.36}$ & $75.72_{\pm 0.43}$ & $67.08_{\pm 0.22}$  \\
                          & \multicolumn{1}{l|}{TENT \cite{TENT}}   & $87.07_{\pm 0.87}$ & $86.76_{\pm 0.52}$ & $96.92_{\pm 0.20}$ & $78.83_{\pm 1.97}$ & $87.40_{\pm 0.69}$ & $64.53_{\pm 0.21}$ & $55.15_{\pm 0.59}$ & $74.68_{\pm 0.38}$ & $76.19_{\pm 0.14}$ & $67.64_{\pm 0.19}$  \\
                          & \multicolumn{1}{l|}{TSD \cite{TSD}}    & $\underline{90.83}_{\pm 0.89}$ & $\underline{89.94}_{\pm 0.27}$ & $\underline{97.56}_{\pm 0.27}$ & $\underline{81.56}_{\pm 1.52}$ & $\underline{89.97}_{\pm 0.18}$ & $65.33_{\pm 0.30}$ & $56.99_{\pm 0.34}$ & $76.12_{\pm 0.36}$ & $76.52_{\pm 0.24}$ & $68.74_{\pm 0.14}$  \\
                          & \multicolumn{1}{l|}{TEA \cite{TEA}}    & $88.51_{\pm 0.43}$ & $87.75_{\pm 0.51}$ & $97.17_{\pm 0.24}$ & $81.45_{\pm 1.02}$ & $88.72_{\pm 0.26}$ & $\underline{65.62}_{\pm 0.45}$ & $\underline{57.55}_{\pm 0.64}$ & $\underline{75.80}_{\pm 0.39}$ & $\underline{76.85}_{\pm 0.51}$ & $\underline{68.95}_{\pm 0.25}$  \\
                          & \multicolumn{1}{l|}{TIPI \cite{TIPI}}   & $87.62_{\pm 0.73}$ & $86.97_{\pm 0.96}$ & $97.01_{\pm 0.11}$ & $79.66_{\pm 1.50}$ & $87.81_{\pm 0.56}$ & $64.92_{\pm 0.29}$ & $56.47_{\pm 0.37}$ & $75.31_{\pm 0.27}$ & $76.74_{\pm 0.37}$ & $68.36_{\pm 0.04}$  \\
                          \rowcolor[HTML]{FADADD}
                          & \multicolumn{1}{l|}{S-FRET}   & ${89.90}_{\pm 0.60}$ & ${89.00}_{\pm 0.48}$ & ${97.46}_{\pm 0.32}$ & ${77.45}_{\pm 0.91}$ & ${88.45}_{\pm 0.19}$ & ${64.42}_{\pm 0.20}$ & ${55.06}_{\pm 0.34}$ & ${75.56}_{\pm 0.24}$ & ${76.41}_{\pm 0.14}$ & ${67.86}_{\pm 0.13}$\\
                          \rowcolor[HTML]{FADADD}
                          & G-FRET                        & $\textbf{90.99}_{\pm 0.87}$ & $\textbf{90.00}_{\pm 0.25}$ & $\textbf{97.59}_{\pm 0.29}$ & $\textbf{82.32}_{\pm 2.59}$ & $\textbf{90.23}_{\pm 0.53}$ & $\textbf{66.12}_{\pm 0.41}$ & $\textbf{57.63}_{\pm 0.39}$ & $\textbf{76.46}_{\pm 0.34}$ & $\textbf{77.19}_{\pm 0.23}$ & $\textbf{69.35}_{\pm 0.20}$  \\ \bottomrule
\end{tabular}
}
\caption{Accuracy comparison of different TTA methods on PACS and OfficeHome datasets based on ResNet-18 and ResNet-50 backbones, including mean and standard deviation of 5 runs with different random seeds. The best results are highlighted in \textbf{boldface}, and the second ones are \underline{underlined}.}
\label{tab:DG}
\end{center}
\end{table*}

\section{Experiments and Analysis}
In this section, we conduct extensive experiments to explore the following key questions:  
(1) Is Test-Time Adaptation (TTA) based on feature redundancy elimination truly effective? (\cref{sebsec:Comparison with State-of-the-art Methods})  
(2) Do the underlying mechanisms of S-FRET and G-FRET align with our theoretical assumptions? (\cref{subsec:Analysis of Underlying Mechanisms}) 
(3) Additionally, we perform ablation studies and evaluate the scalability of our method in more complex scenarios. \cref{subsec:Ablation and Extension Study})


\subsection{Experimental Setup}
We evaluate the adaptation performance on two main tasks: image domain adaptation and image corruption adaptation. Following previous studies, for domain adaptation, we use the PACS \cite{PACS} dataset and the OfficeHome \cite{OfficeHome} dataset. For image corruption adaptation, we utilize the CIFAR10-C , CIFAR100-C and ImageNet-C \cite{hendrycks2019benchmarking} datasets. To implement label shifts, we adjust the CIFAR-100-C datasets to follow a long-tail distribution \cite{du2024probabilistic}, denoted as CIFAR-100-C-LT. The comparison methods we employ include: BN \cite{BN}, TENT \cite{TENT} EATA \cite{EATA}, SAR \cite{SAR}, TSD \cite{TSD}, TIPI \cite{TIPI}, and TEA \cite{TEA}. Backbone networks include ResNet-18 and ResNet-50 \cite{he2016deep}. To ensure fairness, we report mean and standard deviation of 5 runs with different random seeds and independently perform hyperparameter tuning for all methods to achieve the highest accuracy as their final results. Refer to \cref{sec:Experimental details} for more implement information. For further experimental results and analysis, please see \cref{sec:rationale}.

\subsection{Comparison with State-of-the-art Methods}
\label{sebsec:Comparison with State-of-the-art Methods}
\subsubsection{Domain Generalization}
\label{sec:Domain Generalization}
\cref{tab:DG} presents the accuracy comparison of S-FRET and G-FRET with other state-of-the-art TTA methods on the PACS and OfficeHome datasets based on ResNet-18 and ResNet-50 backbones.The table shows that S-FRET, through its simple yet effective optimization, attains competitive performance comparable to most existing TTA methods, while G-FRET achieves superior performance across both datasets and backbone types. Specifically, on the PACS dataset, G-FRET improves upon the baseline non-adaptive source model by 6.67\% and 5.45\% with ResNet-18 and ResNet-50, respectively. For OfficeHome, the improvements are 1.80\% and 2.34\%, respectively. Compared with state-of-the-art TTA methods, G-FRET achieves the highest or second-highest accuracy overall, particularly when using the ResNet-50 as backbone, G-FRET consistently achieves the highest accuracy across all domains. In contrast, other methods fail to achieve consistently high accuracy across different domains and backbones. This is caused by G-FRET's ability to consider both feature discriminability and redundancy.

\subsubsection{Image Corruption}
\label{sec:Image Corruption}
\cref{tab:IC} shows the accuracy comparisons of S-FRET and G-FRET and other state-of-the-art methods for image corruption on CIFAR-10-C , CIFAR-100-C, and ImageNet-C at damage level of 5. We designed two experimental settings: continuous adaptation and independent adaptation. For CIFAR-10/100-C, the model undergoes continuous adaptation, where all 15 corruption types are applied sequentially. For ImageNet-C, each corruption type is evaluated independently on the adapted model. The results demonstrate that S-FRET significantly outperforms the baseline non-adaptive source model and achieves average performance increases of 22.84\% on CIFAR-10-C, 17.52\% on CIFAR-100-C and 14.4\% on ImageNet-C. It can be observed that G-FRET consistently outperforms S-FRET. This is because S-FRET only considers feature redundancy without incorporating label distribution information, whereas G-FRET accounts for both feature redundancy and discriminability. For continuous adaptation, compared to other methods that do not explicitly minimize feature redundancy, their performance is similar to S-FRET but inferior to G-FRET. This indicates that both feature redundancy reduction and traditional class discriminative information contribute to continuous TTA performance, and their combination further enhances adaptation. This finding is further validated in the subsequent ablation studies.

\begin{table}[h]
\centering
\small
\resizebox{\columnwidth}{!}{
\begin{tabular}{@{}l|ccc|ccc|ccc@{}}
\toprule
\multirow{2}{*}{Method} & \multicolumn{3}{c|}{CIFAR-10-C}                  & \multicolumn{3}{c|}{CIFAR-100-C}  & \multicolumn{3}{c}{ImageNet-C}                \\ \cmidrule(l){2-10} 
                        & 1-7        & 8-15       & 1-15       & 1-7        & 8-15       & 1-15   & 1-7        & 8-15       & 1-15     \\ \midrule
Source \cite{he2016deep}                     & 38.46          & 61.59          & 50.80          & 23.59          & 37.49          & 31.01        & 7.74 & 20.79 & 14.70 \\
BN \cite{BN}                      & 70.47          & 76.43          & 73.65          & 45.70          & 50.57          & 48.30       &17.84 & 36.12 & 27.59   \\
TENT \cite{TENT}                     & 73.06          & 79.25          & 76.36          & \underline{47.77}    & \underline{53.44}    & \underline{50.79}   &27.05 & 42.69 & 35.39 \\
EATA \cite{EATA}                    & 70.60          & 76.42          & 73.70          & 47.64          & 53.40          & 50.71       &\textbf{31.23} & \textbf{46.15} & \textbf{39.19}   \\
SAR \cite{SAR}                    & 70.61          & 76.50          & 73.75          & 46.66          & 51.21          & 49.09      &\underline{30.37} & \underline{45.70} & \underline{38.55}    \\
TIPI \cite{TIPI}                     & \textbf{73.76} & \underline{80.21}    & \underline{77.21}    & 47.50          & 52.06          & 49.93    &27.78 & 42.69 & 35.73      \\
TEA \cite{TEA}                     & 71.70          & 77.47          & 74.78          & 46.07          & 50.64          & 48.51        &22.85 & 40.72 & 32.38  \\
TSD \cite{TSD}                    & \underline{72.53}          & 77.96          & 75.43          & 45.87          & 51.03          & 48.62      &20.93 & 38.15 & 30.11    \\
\rowcolor[HTML]{FADADD}
S-FRET                    & 71.60 & 77.28 & 74.63 & 46.31 & 50.46 & 48.53 &19.59 & 37.43 & 29.10\\
\rowcolor[HTML]{FADADD}
G-FRET                    & \textbf{73.76} & \textbf{80.32} & \textbf{77.26} & \textbf{48.23} & \textbf{53.55} & \textbf{51.06} &28.60 & 43.96 & 36.79 \\ \bottomrule
\end{tabular}
}
\caption{Accuracy comparison of different TTA methods on CIFAR-10/100-C and ImageNet-C datasets at damage level of 5, with 15 types of damage applied two experimental settings: continuous adaptation and independent adaptation.}
\label{tab:IC}
\end{table}

\subsubsection{Efficiency Comparison}
In addition to accuracy, we also evaluate the runtime efficiency of different methods to assess the computational performance of S-FRET and G-FRET. As shown in \cref{tab:Eff}, "Training" indicates whether the method involves retraining. ERM represents the source model without adaptation, while the BN method directly updates the batch normalization (BN) parameters, eliminating the need for retraining. Consequently, BN is the most computationally efficient method. However, as reported in \cref{tab:DG,tab:IC}, BN consistently exhibits the lowest accuracy. Among the traning methods, S-FRET achieves the lowest runtime across different backbones, whereas G-FRET operates at an average runtime compared to other baselines. These results indicate that S-FRET, as a simple yet effective approach, provides a competitive accuracy-efficiency trade-off. Meanwhile, G-FRET, despite its moderate computational cost, achieves the highest accuracy, demonstrating its effectiveness in balancing efficiency and adaptation performance.

\begin{table}[h]
\centering
\small
\resizebox{\columnwidth}{!}{
\begin{tabular}{@{}c|cc|*{6}{c}>{\columncolor[HTML]{FADADD}}c>{\columncolor[HTML]{FADADD}}c@{}}
\toprule
\multicolumn{11}{c}{Cost Time (s)} \\ \midrule
Backbone & ERM & BN & TENT & EATA & SAR & TSD & TIPI & TEA  & S-FRET & G-FRET \\ \midrule
Traning & $\times$ & $\times$ & $\checkmark$ & $\checkmark$ & $\checkmark$ & $\checkmark$ & $\checkmark$ & $\checkmark$ & $\checkmark$ & $\checkmark$ \\ \midrule
ResNet-18    & 39 & 36 & \underline{48} & 56 & 74 & 114 & 162 & 658 & \textbf{46} & 118 \\
ResNet-50    & 64 & 66 & 113 & \underline{109} & 185 & 153 & 264 & 1376 & \textbf{102} & 193 \\
VIT-B/16    & 141 & 162 & \underline{210} & 220 & 345 & 295 & 281 & 3181 & \textbf{185} & 310 \\ \midrule \rowcolor{blue!10}
AVG  & 81 & 88 & \underline{124} & 128 & 201 & 187 & 236 & 1738 & \textbf{111} & 207 \\
\bottomrule
\end{tabular}
}
\caption{Cost time comparison of TTA methods on ImageNet-C (Gaussian Noise) across multiple backbone networks}
\label{tab:Eff}
\end{table}

\subsection{Analysis of Underlying Mechanisms}
\label{subsec:Analysis of Underlying Mechanisms}
\subsubsection{Relation between Feature Redundancy Elimination and Generalizability Enhancement}
\begin{figure}[t]
\centering
\includegraphics[width=\columnwidth]{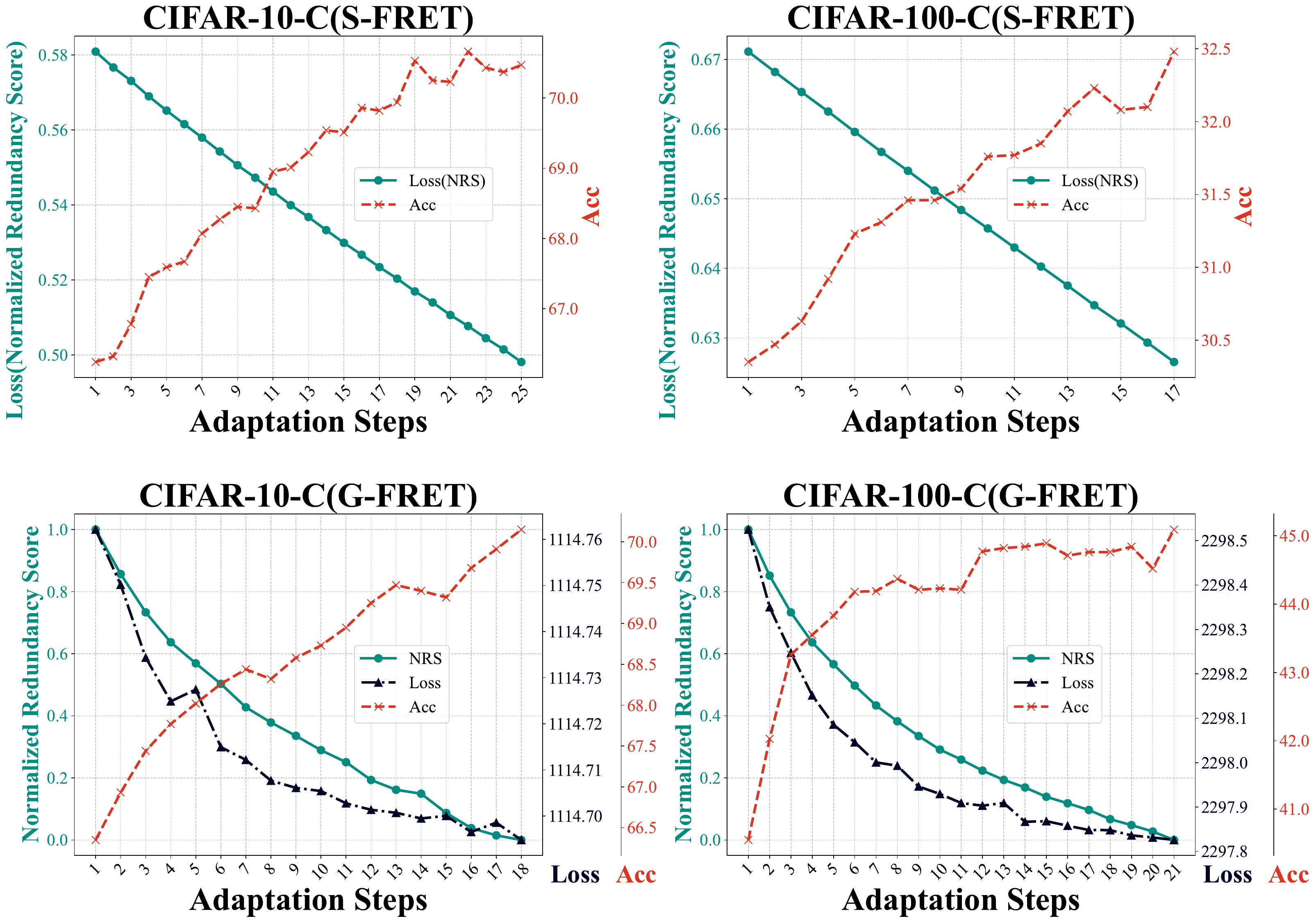} 
\caption{The relationship of the reduction of redundancy and the enhancement of generalizability attained by S-FRET and G-FRET on CIFAR-10-C and CIFAR-100-C datasets. NRS: Normalized Redundancy Scores.}
\label{fig:fig3}
\end{figure}

The first mechanism to be validated is whether there exists a genuine relationship between feature redundancy and generalizability enhancement. We track the evolution of normalized redundancy scores (NRS), loss, and accuracy across increasing adaptation steps. As shown in \cref{fig:fig3}, S-FRET achieves a consistent reduction in feature redundancy as adaptation progresses, accompanied by stable accuracy improvement, demonstrating the detrimental effect of redundancy on generalization and the efficacy of redundancy elimination. For G-FRET, as the number of adaptation steps increases, there is also a consistent reduction in the redundancy of the extracted embedded features, accompanied by a significant decrease in loss and a notable improvement in accuracy. In summary, both S-FRET and G-FRET has been demonstrated to be highly effective in minimizing redundancy, and redundancy elimination is strongly associated with enhancement of model generalizability.

\subsubsection{Label Shift}

\begin{table}[h]
\centering
\small
\resizebox{\columnwidth}{!}{
\begin{tabular}{@{}c|*{8}{c}>{\columncolor[HTML]{FADADD}}c>{\columncolor[HTML]{FADADD}}c@{}}
\toprule
\multicolumn{11}{c}{CIFAR-100-C-LT} \\ \midrule
Imbalance Factor & ERM & BN & SAR & EATA & TENT & TSD & TIPI & TEA  & S-FRET & G-FRET \\ \midrule
1    & 31.01 & 48.30 & 49.09 & 50.71 &\underline{50.79} & 48.62 & 48.51 & 49.93 & 48.53 & \textbf{51.06} \\
10    & 30.39 & 47.14 & 47.69 & 47.86 & \underline{50.12} & 47.74 & 47.50 & 49.54 & 47.47 & \textbf{50.23} \\
100    & 30.13 & 46.07 & 48.21 & 46.08 & \underline{49.80} & 46.95 & 47.13 & 49.51 & 46.62 & \textbf{50.27} \\ \midrule \rowcolor{blue!10}
AVG  &30.51 & 47.17 & 48.33 & 48.21 & \underline{50.24} & 47.77 & 47.71 & 49.66 & 47.54 & \textbf{50.52} \\
\bottomrule
\end{tabular}
}
\caption{Performance comparison on CIFAR-100-C-LT with different imbalance factors based on ResNet-18.}
\label{tab:LS2}
\end{table}


Is S-FRET truly affected by label shift? \cref{tab:LS2} presents the performance of S-FRET and G-FRET under test-time label shifts with long-tailed distributions. Here, the imbalance factor denotes the ratio between the most and least frequent class samples—higher values indicate more skewed class distributions. As shown in \cref{tab:LS2}, S-FRET achieves competitive performance on CIFAR-100-C under balanced class distributions. However, its accuracy degrades significantly as the imbalance factor increases, with 1.91\% drops on CIFAR-100-C, aligning with the limitations of conventional TTA methods. In contrast, G-FRET delivers more robust performance across all imbalance factors. This highlights G-FRET’s unique capability to address label shift in real-world scenarios through contrastive learning.

\subsubsection{Discriminability Analysis by tSNE}
\begin{figure}[t]
\centering
\includegraphics[width=\columnwidth]{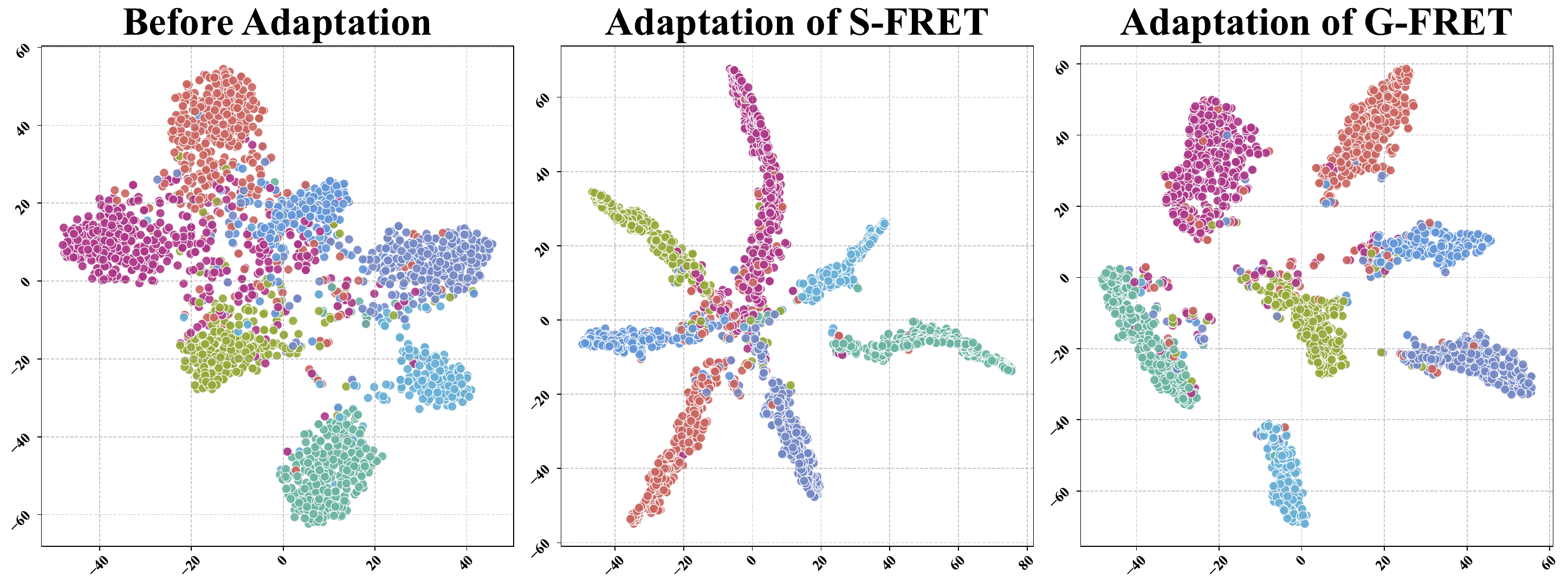} 
\caption{Discriminability visualization of embedded feature before and after adaptation by S-FRET and G-FRET respectively.}
\label{fig:fig5}
\end{figure}

Does G-FRET truly improve discriminability compared to S-FRET? \cref{fig:fig5} shows the t-SNE \cite{Maaten2008VisualizingDU} visualization of embedded features before and after adaptation by S-FRET and G-FRET. It is evident that before adaptation, data points from different classes are intermixed without clear separation. After adaptation by S-FRET, although class separation increases, the overlapping regions between classes still lack distinct boundaries. In contrast, G-FRET achieves more distinct cluster separation and sharper decision boundaries, demonstrating its superior capability in learning discriminative features under distribution shifts.

\subsection{Ablation and Extension Study}
\label{subsec:Ablation and Extension Study}
\subsubsection{Effects of Components in G-FRET}
\cref{tab:Ablation} presents the impact of the four components of G-FRET on model performance. (1) Compared to the source baseline, each individual component contributes positively to the model's performance, with respective gains of 0.30\%, 1.11\%, 1.15\%, and 0.06\%. Notably, the redundancy component in the representation layer, \( \mathcal{L}_R(R) \), and the attention component in the prediction layer, \( \mathcal{L}_P(A) \), show the most significant impact, with \( \mathcal{L}_R(R) \) achieving a 1.15\% improvement. This highlights the importance of redundancy reduction for TTA. (2) The ablation study further confirms the significance of each component, indicating that their removal negatively impacts model performance. (3) Either the redundant or attention components can individually improves model performance, with respective gains of 1.28\% and 1.17\%. When the two components are jointly considered, they yield an even greater improvement. (4) Overall, the experimental results further validate the necessity of considering feature redundancy during the TTA process.

\begin{table}[t]
\centering
\small
\resizebox{\columnwidth}{!}{
    \begin{tabular}{cccllllll}
    \toprule
    \multicolumn{9}{c}{OfficeHome}                                                                                                                                                                                                                           \\ \midrule
    \multicolumn{3}{c|}{Components}                                                       & \multicolumn{1}{c}{A} & \multicolumn{1}{c}{C} & \multicolumn{1}{c}{P} & \multicolumn{1}{c|}{R}              & \multicolumn{1}{c}{AVG} & \multicolumn{1}{c}{$\uparrow$} \\ \midrule
    \rowcolor[HTML]{EFEFEF}
    \multicolumn{3}{c|}{Source}                        & 56.45                 & 48.02                 & 71.34                 & \multicolumn{1}{l|}{72.23}          & 62.01                   & 0.00                            \\ \midrule
    \multicolumn{1}{c|}{\multirow{4}{*}{Individual}} & \multicolumn{1}{c|}{\multirow{2}{*}{$\mathcal{L}_R$}} & \multicolumn{1}{c|}{$A$}  & 55.83                 & 49.67                 & 70.83                 & \multicolumn{1}{l|}{72.89}          & 62.31                   & 0.30                            \\
    \multicolumn{1}{c|}{}       & \multicolumn{1}{c|}{}                    & \multicolumn{1}{c|}{$R$}   & 56.98                 & 50.42                 & 72.33                 & \multicolumn{1}{l|}{72.73}          & 63.12                   & 1.11                            \\
    \multicolumn{1}{c|}{}       & \multicolumn{1}{c|}{\multirow{2}{*}{$\mathcal{L}_P$}}        & \multicolumn{1}{c|}{$A$}   & 56.73                 & 50.77                 & 72.13                 & \multicolumn{1}{l|}{72.99}          & 63.16                   & 1.15                            \\
    \multicolumn{1}{c|}{}       & \multicolumn{1}{c|}{}        & \multicolumn{1}{c|}{$R$}   & 55.83                 & 49.30                 & 70.62                 & \multicolumn{1}{l|}{72.51}          & 62.07                   & 0.06                            \\ \midrule
    \multicolumn{1}{c|}{\multirow{4}{*}{Ablation}}  & \multicolumn{1}{c|}{\multirow{2}{*}{w/o $\mathcal{L}_R$}}  & \multicolumn{1}{c|}{$A$}   & 57.72                 & 51.18                 & 72.99                 & \multicolumn{1}{l|}{73.08}          & 63.74                   & 1.73                            \\
    \multicolumn{1}{c|}{}              & \multicolumn{1}{c|}{}              & \multicolumn{1}{c|}{$R$}   & 56.70                 & 51.16                 & 72.07                 & \multicolumn{1}{l|}{73.19} & 63.28                   & 1.27                            \\
    \multicolumn{1}{c|}{}              & \multicolumn{1}{c|}{\multirow{2}{*}{w/o $\mathcal{L}_P$}}              & \multicolumn{1}{c|}{$A$}   & 57.11                 & 50.52                 & 72.40                 & \multicolumn{1}{l|}{72.94}          & 63.24                   & 1.23                            \\
    \multicolumn{1}{c|}{}                 & \multicolumn{1}{c|}{}           & \multicolumn{1}{c|}{$R$}   & 57.77                 &51.40       & 72.94                 & \multicolumn{1}{l|}{72.96}          & 63.77                   & 1.76                            \\ \midrule
    \multicolumn{1}{c|}{Attention}                   & \multicolumn{2}{c|}{$\mathcal{L}_R(A)$+$\mathcal{L}_P(A)$} & 56.70                 & 50.97                 & 72.36                 & \multicolumn{1}{l|}{73.12}          & 63.29                   & 1.28                            \\ \midrule
    \multicolumn{1}{c|}{Redundant}                   & \multicolumn{2}{c|}{$\mathcal{L}_R(R)$+$\mathcal{L}_P(R)$} & 57.03                 & 50.49                 & 72.27                 & \multicolumn{1}{l|}{72.94}          & 63.18                   & 1.17                            \\ \midrule
    \rowcolor{blue!10}
    \multicolumn{1}{c|}{All}                         & \multicolumn{2}{c|}{All}     & 57.49        & 51.28                 & 73.04        & \multicolumn{1}{l|}{74.40}          & \textbf{63.81}          & \textbf{1.80}                  \\ 
    \bottomrule
    \end{tabular}
    } 
\caption{Effects of components in G-FRET, test on the OfficeHome dataset with ResNet-18 backbone. A: Attention part, R: Redundancy Part.}
\label{tab:Ablation}
\end{table}

\subsubsection{Parameter Sensitivity}
\begin{figure}[t]
\centering
\includegraphics[width=0.8\columnwidth]{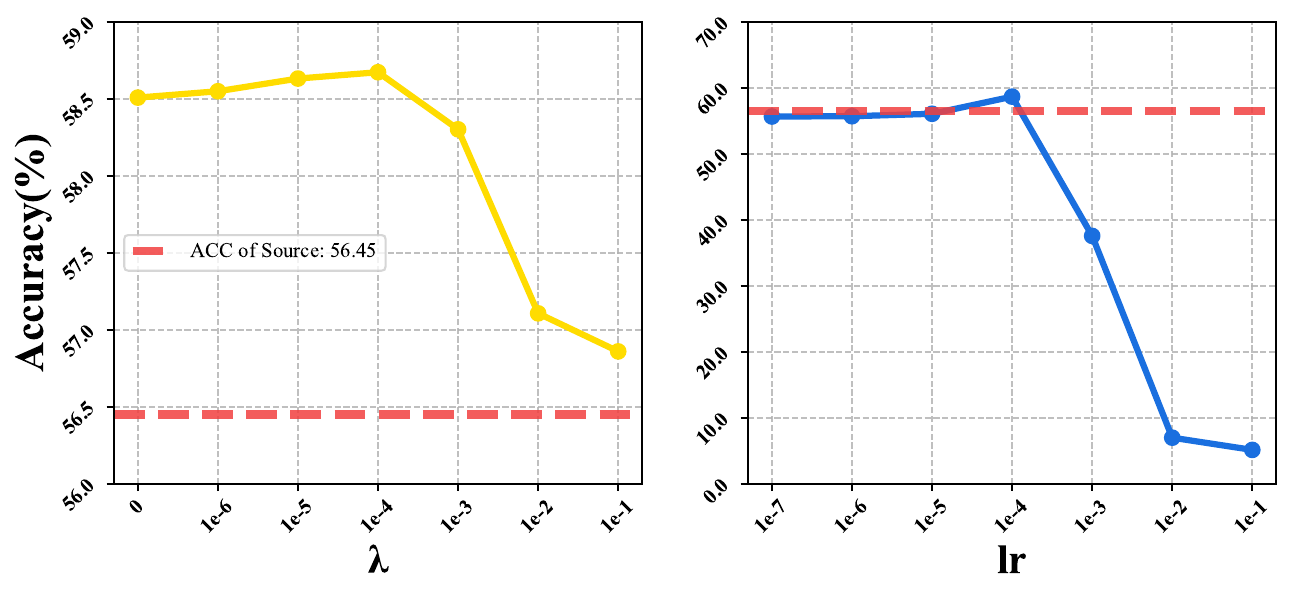} 
\caption{Parameter sensitivity analysis G-FRET on the OfficeHome dataset with ResNet-18 backbone in domain Art.}
\label{fig:fig4}
\end{figure}

Our method G-FRET involves one hyperparameter: the trade-off parameter in \cref{eq:L}, $\lambda$. As shown in \cref{fig:fig4}, G-FRET outperforms the source model across a wide range of values for $\lambda$. We also investigate the impact of the learning rate (lr) on the experimental results. The best performance is achieved with a learning rate of around $1 \times 10^{-4}$; higher learning rates tend to destabilize the training process and reduce the effectiveness of model, which is also observed in other TTA methods \cite{TSD,TENT,TEA,TIPI,SAR,EATA}.

\subsubsection{Scalability on complex/large dataset and different backbones.}
\label{sec:Scalability Experiment}
We validated our method on more complex and larger datasets, as well as on a VIT \cite{VIT} backbone, to ensure that our approach can robustly improve performance and is not dependent on any specific network architecture. \cref{tab:Scalability} demonstrates that the proposed methods generally enhances baseline performance across different backbones. Even on large and complex datasets such as ImageNet-C\cite{hendrycks2019benchmarking}, our method demonstrates significant improvements across different backbones, with S-FRET achieving performance gains of 14.4\%, 16.5\%, 20.53\% and G-FRET attaining 22.09\%, 24.69\%, 22.67\%, respectively.

\begin{table}[t]
\centering
\small
\resizebox{\columnwidth}{!}{
    \begin{tabular}{lccc}
    \toprule
    Method & VLCS\cite{VLCS} & DomainNet\cite{DomainNet} & ImageNet-C\cite{hendrycks2019benchmarking} \\ \midrule
    ResNet-18 \cite{he2016deep} & 73.67 & 39.13 & 14.70 \\
    +S-FRET & 73.90 & 39.26 & 29.10\\
    +G-FRET & \textbf{75.73} & \textbf{40.22} & \textbf{36.79}\\ \midrule
    ResNet-50 \cite{he2016deep} & 74.90 & 43.58 & 18.15 \\
    +S-FRET & 75.48 & 43.69 & 34.65\\
    +G-FRET & \textbf{76.42} & \textbf{44.22} & \textbf{42.84}\\ \midrule
    VIT-B/16 \cite{VIT} & 76.44 & 49.98 & 39.83 \\
    +S-FRET & 77.18 & 51.22 & 60.36\\
    +G-FRET & \textbf{78.56} & \textbf{51.50} & \textbf{62.50}\\ \midrule
    \end{tabular}
    } 
\caption{Scalability on complex/large dataset and different backbones.}
\label{tab:Scalability}
\end{table}

\section{Conclusion}
In this work, we propose a novel approach for test-time adaptation (TTA) — Feature Redundancy Elimination for TTA (FRET) — which focuses on reducing feature redundancy during testing to address distribution shifts. This is the first attempt to leverage feature redundancy for TTA. We introduce two methods: S-FRET, which directly minimizes the feature redundancy score, and G-FRET, which optimizes both redundancy elimination and class discriminability by decomposing feature relations into attention and redundancy components. Experiments across various tasks and datasets show the effectiveness of S-FRET and G-FRET, demonstrating the value of feature redundancy elimination in TTA.  
{
    \small
    \bibliographystyle{ieeenat_fullname}
    \bibliography{main}
}


\clearpage
\setcounter{page}{1}
\maketitlesupplementary

\section{Extended Related Work}
\label{sec:Extended Related work}

\subsection{Test-Time Adaptation}

In realistic scenarios, test data often undergoes natural variations or corruptions, resulting in data distribution shifts between the training and test phases. Recently, various Test-Time Adaptation (TTA) approaches have been proposed to adapt pre-trained models during testing  \cite{tan2024heterogeneity,yuan2023robust}. For batch normalization calibration methods, BN \cite{BN} replaces the activation statistics estimated from the training set with those of the test set. For pseudo-labeling methods, TSD \cite{TSD} filters out unreliable features or predictions with high entropy, as lower entropy typically indicates higher accuracy, and it further filters unreliable samples using a consistency filter. For consistency training methods, TIPI \cite{TIPI} identifies input transformations that can simulate domain shifts and uses regularizers, based on derived bounds, to ensure the network remains invariant to such transformations. For clustering-based training methods, TENT \cite{TENT} minimizes the prediction entropy of the model on the target data. EATA \cite{EATA} selects reliable samples to minimize entropy loss during test-time adaptation and uses a Fisher regularizer to stabilize key parameters. The importance of these parameters is estimated from test samples with pseudo labels. SAR \cite{SAR} removes noisy samples with large gradients and encourages model weights to reach a flat minimum, enhancing robustness against remaining noise. Recently, TEA \cite{TEA}, a state-of-the-art TTA approach, introduces an innovative energy-based perspective to mitigate the effects of distribution changes and has shown advantages over current leading approaches.

\subsection{Feature Redundancy Elimination}
Feature redundancy is a key concern in both feature extraction and feature selection \cite{FE_review2,FS_review3}. Feature extraction methods aim to reduce redundancy by transforming the original feature space into a new low-dimensional feature space while retaining as much relevant information as possible. Two typical feature extraction methods are unsupervised method Principal Component Analysis (PCA) \cite{PCA} and supervised method Linear Discriminant Analysis (LDA) \cite{LDA}. The former one performs a linear transformation to create a new feature space where the features are uncorrelated, while the latter one reduces redundancy by identifying feature spaces that best separate different classes by maximizing the between-class dispersion while minimizing within-class dispersion.

Unlike feature extraction, feature selection aims to identify the most representative and non-redundant subset of features from the original feature set \cite{FS_review2}. Feature selection methods generally include three strategies: 1) Filter methods \cite{covert2023learning,brown2012conditional} use statistical measures (e.g., mutual information, Fisher score) to evaluate feature relevance and redundancy. Features with high relevance are deemed more informative for target variables, while redundant features are removed to enhance feature independence. 2) Wrapper methods \cite{ramjee2019efficient} assess feature subsets by training models and evaluating their performance. A common wrapper method is Recursive Feature Elimination (RFE) \cite{chen2007enhanced}, which iteratively removes features and evaluates model performance to identify the optimal subset. 3) Embedded methods \cite{gui2016feature,hou2023adaptive} integrate feature selection within the model training process, such as Lasso regression \cite{tibshirani1996regression}, which penalizes redundant features during training to optimize model parameters and select the most relevant features. Recently, a novel feature selection approach called SOFT \cite{second_order} has been proposed, which combines second-order covariance matrices with first-order data matrices by knowledge contrastive distillation for unsupervised feature selection. 

\section{Some Filter Tricks}
\begin{figure}[t]
\centering
\includegraphics[width=\columnwidth]{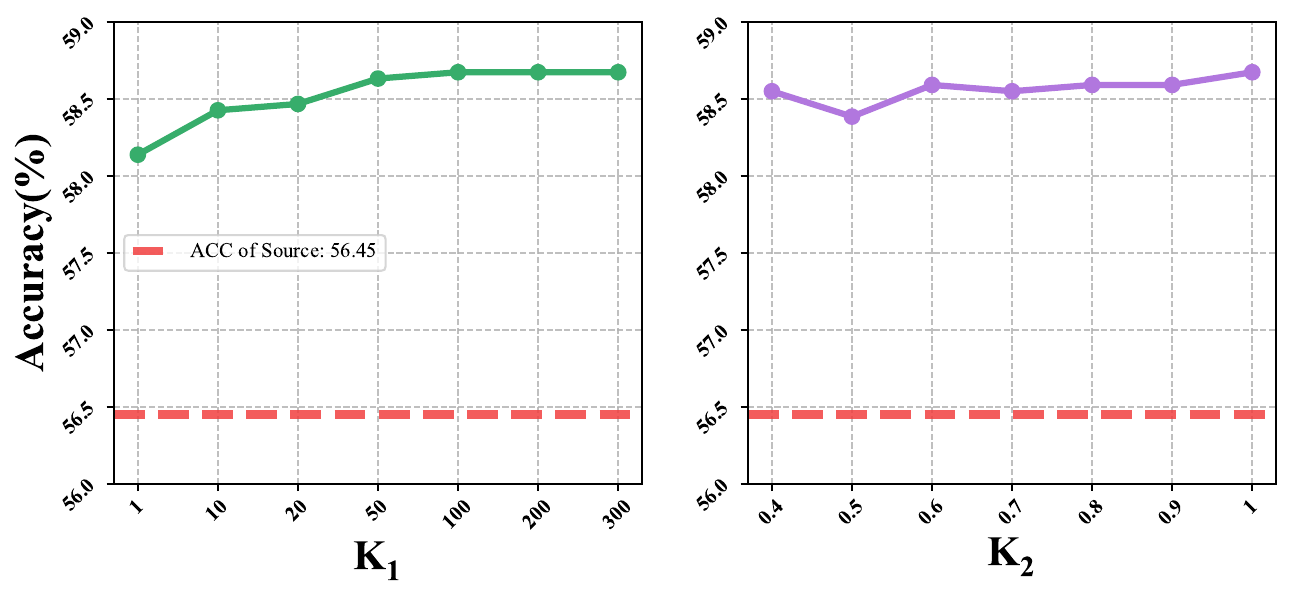} 
\caption{Sensitivity analysis of two G-FRET filtering parameters using ResNet-18 on OfficeHome’s Art domain.}
\label{fig:filter}
\end{figure}
Following the previous works \cite{TSD,TENT,EATA}, to reduce the influence of wrong and noisy results from model $g$, which may create some incorrect computations, we use Shannon entropy \cite{shannon1948mathematical} to filter unreliable instances. Specifically, for each class, we only take the representations with the top-$K_1$ lowest output entropy $H_i$ into computing class centers:

\begin{equation}
H_i = -\sum \sigma(p_i) \log \sigma(p_i)
\end{equation}

Furthermore, we use the similarity between representation and class centers to generate soft pseudo labels $\hat{y}_i$ for the $i$th instance, which can be formulated as:

\begin{equation}
\hat{y}_i = \sigma\left([ \text{sim}(R_{A_i}, c_1), \text{sim}(R_{A_i}, c_2), \ldots, \text{sim}(R_{A_i}, c_C) ]\right)
\end{equation}

Based on this, we only consider the predictions with the top-$K_2$ percent lowest output entropy in a data batch while keeping label consistency between pseudo labels and their predictions, i.e., $\text{argmax}(p_i) = \text{argmax}(\hat{y}_i)$, when computing $L_R$ and $L_P$. We evaluate the effectiveness of these filter tricks on G-FRET. As shown in \cref{fig:filter}, G-FRET exhibits insensitivity to $K_1$, and $K_2$.

\section{Experimental details}
\label{sec:Experimental details}
\subsection{Datasets}
We evaluate the performance of the S-FRET and G-FRET on two main tasks: domain generalization and image corruption generalization. Following previous studies, for domain generalization, we use the PACS \cite{PACS} dataset, consisting of images from seven categories (e.g., objects, animals) across four domains (art paintings, cartoons, photos, and sketches), and the OfficeHome \cite{OfficeHome} dataset, which includes 65 categories (e.g., office items, home objects) from four domains (art, clipart, product, and real-world images). For image corruption generalization, we utilize the CIFAR10-C , CIFAR100-C and ImageNet-C \cite{hendrycks2019benchmarking} datasets, which contain 15 types of corruption at five severity levels. To be consistent with prior research \cite{lee2024aetta}, all experiments are conducted at the highest severity level (level 5). To implement label shifts, we adjust the CIFAR-100-C datasets to follow a long-tail distribution \cite{du2024probabilistic}, denoted as CIFAR-100-C-LT. 

\subsection{Comparison Methods}
The comparison methods we employ include the non-adaptive source model and four baseline methods which are commonly used in several studies: BN \cite{BN}, TENT \cite{TENT}, EATA \cite{EATA}, and SAR \cite{SAR}. Additionally, we adopt three recently proposed state-of-the-art methods: TSD \cite{TSD}, TIPI \cite{TIPI}, and TEA \cite{TEA}.

\subsection{Models and Implementation}
For domain generalization tasks, we use ResNet-18/50 \cite{he2016deep} with batch normalization as the backbone network. These networks are pretrained on data from three domains and then tested on the remaining domain. 

For image corruption, we use ResNet-18 as the backbone network. We train it on the clean versions of the CIFAR-10 and CIFAR-100 datasets, and for ImageNet-C, we leverage pre-trained model from TorchVision. When it comes to CIFAR-10/100-C, we apply all 15 corruption types sequentially to assess continuous adaptation capabilities. For ImageNet-C, we apply each of the 15 corruption types independently to the adapted model.

To ensure fairness, we report mean and standard deviation of 5 runs with different random seeds (0, 1, 2, 3, 4). In addition, we set the batch size as 128 during testing and independently perform hyperparameter tuning for each method to achieve the highest possible accuracy as the final result. All implementations are carried out using PyTorch \cite{pytorch} and executed on a single NVIDIA 4090 GPU for all experiments.

\subsection{Complex/Large Dataset}
\textbf{VLCS} \cite{VLCS} contains four domains: Caltech101, LabelMe, SUN09, and VOC2007, with a total of 10,729 images across 5 classes. The label distribution across the domains in VLCS exhibits substantial variation , which might be a contributing factor to the poor performance of most Test-Time Adaptation methods on this dataset \cite{TSD}.

\noindent\textbf{DomainNet} \cite{DomainNet} is a large-scale dataset used in transfer learning, consisting of six domains: Clipart, Infograph, Painting, Quickdraw, Real, and Sketch. It consists of a total of 586,575 images, with each domain containing 345 classes.

\noindent\textbf{ImageNet-C} \cite{hendrycks2019benchmarking} is significantly larger compared to the CIFAR10-C and CIFAR100-C. CIFAR10-C and CIFAR100-C consist of 50,000 training images and 10,000 test images each, divided into 10 and 100 classes respectively. In contrast, ImageNet-C contains 1,281,167 training images and 50,000 test images, categorized into 1,000 classes. Specifically, ImageNet-C encompasses 15 types of corruption with five levels of severity. In our experiments, we employed the highest corruption level (level 5). For the pre-trained model on ImageNet-C, we utilize the model provided by TorchVision.

\subsection{Experiment Setting Details}
For hyper-parameter selection in Domain Generalization task (\cref{sec:Domain Generalization}), we first identify the optimal parameter set based on the highest accuracy achieved on the default domain (art paintings in PACS and art in OfficeHome). These parameters are then applied to other domains to assess their performance. Specifically, we conduct a search for the learning rate within the range \{1e-7, 5e-7, 1e-6, 5e-6, 1e-5, 5e-5, 1e-4, 5e-4, 1e-3, 5e-3, 1e-2, 5e-2\}. For methods that include an entropy filter component (e.g., TSD, G-FRET), we explore the entropy filter hyperparameter in the set \{1, 5, 10, 15, 20, 50, 100, 200, 300\}. For the G-FRET, we perform hyperparameter tuning for $\lambda$ within the range \{1e-6, 1e-5, 1e-4, 1e-3, 1e-2, 1e-1, 1, 10, 100\} and $K_{2}$ within the values \{0.5, 0.6, 0.7, 0.8, 0.9, 1\}.
 
For the Image Corruption task (\cref{sec:Image Corruption}), each Test-Time Adaptation (TTA) method continuously adapts to 15 types of image corruptions in the specified order for CIFAR-10/100-C: [Gaussian Noise, Shot Noise, Impulse Noise, Defocus Blur, Glass Blur, Motion Blur, Zoom Blur, Snow, Frost, Fog, Brightness, Contrast, Elastic Transformation, Pixelate, JPEG Compression]. However, for ImageNet-C, we adopt a strategy of independently adapting to each of the 15 corruption types separately.
The hyperparameter ranges remain consistent with those utilized in Domain Generalization. The best performance results obtained for each method are selected as the final experimental outcomes.

\section{Additional Experimental Results}
\label{sec:rationale}

\subsection{Detailed Results Across Five Random Seeds}
To ensure the fairness of our evaluation, we conduct experiments using five different random seeds (0, 1, 2, 3, and 4). The detailed results corresponding to each random seed are presented in \cref{tab:random0,tab:random1,tab:random2,tab:random3,tab:random4}, which highlights the robustness and consistency of our proposed methods S-FRET and G-FRET.

\subsection{Detailed Results for Image Corruption}
In this section, we provide a complete listing of comparisons between S-FRET, G-FRET, and other state-of-the-art methods for Image Corruption (\cref{sec:Image Corruption}) on CIFAR-10/100-C and ImageNet-C datasets at damage level of 5, as shown in \cref{tab:CIFAR10,tab:CIFAR100,tab:ImageNet}.

\subsection{Detailed Results for Scalability Experiment}
In the Scalability experiment (\cref{sec:Scalability Experiment}), we validate our methods using larger and more complex datasets including VLCS, DomainNet, and ImageNet-C, as well as on the ViT backbone, to demonstrate that our approach can robustly improve performance across diverse datasets and different backbones.

For VLCS and DomainNet, we employ hyperparameter selection within the same range as the Domain Generalization task. However, unlike the Domain Generalization task, we independently selected hyperparameters for each domain rather than applying the parameters from the default domain to others. 

For ImageNet-C, we adapt the TTA method to each corruption type individually. We select hyperparameters optimized for the default corruption type (Gaussian Noise), and applied these parameters to other corruption types. The detailed results are presented in \cref{tab:VLCS}, \cref{tab:DomainNet} , and \cref{tab:ImageNet2} .

\begin{table*}[h]
\begin{center}
\small
\begin{tabular}{@{}l|l|cccc>{\columncolor{blue!10}}c|cccc>{\columncolor{blue!10}}l@{}}
\toprule
\multirow{2}{*}{Backbone} & \multirow{2}{*}{Method}   & \multicolumn{4}{c}{PACS} & \multicolumn{1}{c}{\multirow{2}{*}{Avg}} & \multicolumn{4}{c}{OfficeHome} & \multicolumn{1}{c}{\multirow{2}{*}{Avg}} \\ \cmidrule(lr){3-6} \cmidrule(lr){8-11}
                          &                           & A              & C              & P              & S              & \multicolumn{1}{c}{}           & A              & C              & P              & R              & \multicolumn{1}{c}{}                     \\ \midrule
\multirow{9}{*}{ResNet-18} & \multicolumn{1}{l|}{Source \cite{he2016deep}} &78.37 & 77.39 & 95.03 & 76.58 & 81.84 & 56.45 & 48.02 & 71.34 & 72.23 & 62.01 \\ 
& \multicolumn{1}{l|}{BN \cite{BN}}     &80.91 & 80.80 & 95.09 & 73.84 & 82.66 & 55.62 & 49.32 & 70.60 & 72.66 & 62.05 \\ 
& \multicolumn{1}{l|}{SAR \cite{SAR}}   &83.30 & 82.17 & 95.09 & 79.69 & 85.06 & 57.15 & 50.31 & 70.24 & 72.34 & 62.51 \\ 
& \multicolumn{1}{l|}{EATA \cite{EATA}}   &82.71 & 81.53 & 94.91 & 74.19 & 83.34 & 56.41 & 49.62 & 71.66 & 72.27 & 62.49 \\ 
& \multicolumn{1}{l|}{TENT \cite{TENT}}    &82.76 & 82.68 & 95.33 & 78.19 & 84.74 & 56.94 & \underline{50.65} & 71.86 & 72.92 & 63.09 \\ 
& \multicolumn{1}{l|}{TSD \cite{TSD}}   &\textbf{86.96} & \underline{86.73} & \underline{96.41} & 81.22 & \underline{87.83} & \underline{58.06} & 49.81 & 71.37 & 70.67 & 62.47 \\ 
& \multicolumn{1}{l|}{TEA \cite{TEA}}    &86.47 & 85.79 & 95.69 & 80.81 & 87.19 & \textbf{58.63} & 50.56 & 71.95 & 72.92 & \underline{63.52} \\ 
& \multicolumn{1}{l|}{TIPI \cite{TIPI}}    &85.50 & 84.90 & 96.05 & \textbf{83.13} & 87.39 & 57.03 & 50.61 & \underline{72.07} & \textbf{73.28} & 63.25 \\ 
\rowcolor[HTML]{FADADD}
& \multicolumn{1}{l|}{S-FRET} &
86.28 & 86.69 & 96.35 & 74.22 & 85.88 & 56.20 & 50.08 & 71.57 & 72.64 & 62.62\\
\rowcolor[HTML]{FADADD}
& \multicolumn{1}{l|}{G-FRET}   &\underline{86.82} & \textbf{87.03} & \textbf{96.65} & \underline{81.29} & \textbf{87.95} & 57.73 & \textbf{51.36} & \textbf{73.10} & \underline{72.99} & \textbf{63.79} \\ 

 \midrule 
\multirow{9}{*}{ResNet-50} & \multicolumn{1}{l|}{Source \cite{he2016deep}
}&83.89 & 81.02 & 96.17 & 78.04 & 84.78 & 64.85 & 52.26 & 75.04 & 75.88 & 67.01 \\ 
& \multicolumn{1}{l|}{BN \cite{BN}
}     &85.50 & 85.62 & 96.77 & 72.05 & 84.99 & 63.54 & 52.71 & 73.89 & 75.05 & 66.30 \\ 
& \multicolumn{1}{l|}{SAR \cite{SAR}
}   &85.55 & 85.62 & 96.77 & 75.24 & 85.79 & 64.77 & 55.92 & 75.24 & 75.81 & 67.94 \\ 
& \multicolumn{1}{l|}{EATA \cite{EATA}
}   &84.67 & 85.20 & 96.35 & 72.36 & 84.64 & 63.91 & 54.04 & 74.72 & 75.51 & 67.05 \\ 
& \multicolumn{1}{l|}{TENT \cite{TENT}
}    &88.09 & 87.33 & 97.19 & 79.69 & 88.07 & 64.61 & 54.80 & 75.06 & 76.20 & 67.67 \\ 
& \multicolumn{1}{l|}{TSD \cite{TSD}
}   &\underline{90.43} & \underline{89.89} & \textbf{97.84} & \underline{81.80} & \underline{89.99} & 65.27 & 56.77 & \underline{76.19} & 76.41 & 68.66 \\ 
& \multicolumn{1}{l|}{TEA \cite{TEA}
}    &88.09 & 87.88 & 97.49 & 81.39 & 88.71 & \underline{66.25} & \textbf{57.50} & 75.20 & 76.68 & \underline{68.91} \\ 
& \multicolumn{1}{l|}{TIPI \cite{TIPI}
}    &88.18 & 87.93 & 97.13 & 78.80 & 88.01 & 64.73 & 56.24 & 75.47 & \underline{77.00} & 68.36 \\ 
\rowcolor[HTML]{FADADD}
& \multicolumn{1}{l|}{S-FRET}   &
89.99 & 89.51 & \textbf{97.84} & 76.30 & 88.41 & 64.15 & 54.50 & 75.74 & 76.25 & 67.66\\
\rowcolor[HTML]{FADADD}
& \multicolumn{1}{l|}{G-FRET}   &\textbf{90.72} & \textbf{90.15} & \textbf{97.84} & \textbf{82.29} & \textbf{90.25} & \textbf{66.42} & \underline{57.11} & \textbf{76.21} & \textbf{77.35} & \textbf{69.27} \\ 
\bottomrule
\end{tabular}
 \caption{At random seed 0, the accuracy comparison of different TTA methods on PACS and OfficeHome datasets based on ResNet-18 and ResNet-50 backbones. The best results are highlighted in \textbf{boldface}, and the second ones are \underline{underlined}.}
\label{tab:random0}
\end{center}
\end{table*}

\begin{table*}[h]
\begin{center}
\small
\begin{tabular}{@{}l|l|cccc>{\columncolor{blue!10}}c|cccc>{\columncolor{blue!10}}l@{}}
\toprule
\multirow{2}{*}{Backbone} & \multirow{2}{*}{Method}   & \multicolumn{4}{c}{PACS} & \multicolumn{1}{c}{\multirow{2}{*}{Avg}} & \multicolumn{4}{c}{OfficeHome} & \multicolumn{1}{c}{\multirow{2}{*}{Avg}} \\ \cmidrule(lr){3-6} \cmidrule(lr){8-11}
                          &                           & A              & C              & P              & S              & \multicolumn{1}{c}{}           & A              & C              & P              & R              & \multicolumn{1}{c}{}                     \\ \midrule
\multirow{9}{*}{ResNet-18} & \multicolumn{1}{l|}{Source \cite{he2016deep}
} &78.37 & 77.39 & 95.03 & 76.58 & 81.84 & 56.45 & 48.02 & 71.34 & 72.23 & 62.01 \\ 
& \multicolumn{1}{l|}{BN \cite{BN}
}     &81.10 & 80.59 & 95.33 & 73.84 & 82.71 & 55.83 & 48.80 & 70.78 & 72.21 & 61.90 \\ 
& \multicolumn{1}{l|}{SAR \cite{SAR}
}   &84.33 & 80.03 & 95.33 & 79.59 & 84.82 & 56.90 & 50.22 & 70.06 & 72.89 & 62.52 \\ 
& \multicolumn{1}{l|}{EATA \cite{EATA}
}   &82.96 & 82.47 & 95.45 & 73.00 & 83.47 & 57.03 & 50.13 & 71.03 & \underline{72.96} & 62.79 \\ 
& \multicolumn{1}{l|}{TENT \cite{TENT}
}    &84.91 & 81.19 & 95.57 & \underline{82.82} & 86.12 & \underline{57.77} & 50.22 & \underline{72.56} & 72.62 & 63.29 \\ 
& \multicolumn{1}{l|}{TSD \cite{TSD}
}   &86.62 & 86.39 & 95.87 & 81.50 & 87.59 & 57.52 & 49.26 & 72.18 & 71.20 & 62.54 \\ 
& \multicolumn{1}{l|}{TEA \cite{TEA}
}    &\textbf{87.79} & 86.60 & \underline{95.99} & 82.21 & \underline{88.15} & 56.78 & \textbf{50.70} & 72.02 & 72.85 & 63.09 \\ 
& \multicolumn{1}{l|}{TIPI \cite{TIPI}
}    &85.11 & 83.23 & 95.87 & \textbf{85.03} & 87.31 & \textbf{57.81} & 50.19 & \underline{72.56} & \underline{72.96} & \underline{63.38} \\ 
\rowcolor[HTML]{FADADD}
& \multicolumn{1}{l|}{S-FRET}   &
85.84 & \underline{86.69} & 95.93 & 72.84 & 85.33 & 56.53 & 49.51 & 71.71 & 72.53 & 62.57 \\

\rowcolor[HTML]{FADADD}
& \multicolumn{1}{l|}{G-FRET}   &\underline{87.26} & \textbf{86.99} & \textbf{96.59} & 82.74 & \textbf{88.39} & 57.23 & \underline{50.61} & \textbf{73.51} & \textbf{73.17} & \textbf{63.63} \\ 

 \midrule 
\multirow{9}{*}{ResNet-50} & \multicolumn{1}{l|}{Source \cite{he2016deep}
}&83.89 & 81.02 & 96.17 & 78.04 & 84.78 & 64.85 & 52.26 & 75.04 & 75.88 & 67.01 \\ 
& \multicolumn{1}{l|}{BN \cite{BN}
}     &85.01 & 85.88 & 96.65 & 71.88 & 84.85 & 63.00 & 53.54 & 73.60 & 74.96 & 66.27 \\ 
& \multicolumn{1}{l|}{SAR \cite{SAR}
}   &85.01 & 85.88 & 96.65 & 75.92 & 85.86 & 65.06 & 56.31 & 74.50 & 76.45 & 68.08 \\ 
& \multicolumn{1}{l|}{EATA \cite{EATA}
}   &85.35 & 85.41 & 96.71 & 72.23 & 84.92 & 64.81 & 53.88 & 73.80 & 75.60 & 67.02 \\ 
& \multicolumn{1}{l|}{TENT \cite{TENT}
}    &86.23 & 86.95 & 97.01 & 79.77 & 87.49 & 64.81 & 55.58 & 74.75 & 76.38 & 67.88 \\ 
& \multicolumn{1}{l|}{TSD \cite{TSD}
}   &\textbf{89.75} & \underline{89.63} & \textbf{97.49} & \underline{83.76} & \underline{90.16} & \underline{65.72} & 57.14 & \underline{76.39} & 76.18 & 68.86 \\ 
& \multicolumn{1}{l|}{TEA \cite{TEA}
}    &89.16 & 88.05 & 96.89 & 82.54 & 89.16 & 65.22 & \underline{57.69} & 75.67 & \textbf{77.37} & \underline{68.99} \\ 
& \multicolumn{1}{l|}{TIPI \cite{TIPI}
}    &87.16 & 86.99 & 97.07 & 80.99 & 88.05 & 65.39 & 56.11 & 75.69 & 76.27 & 68.36 \\ 
\rowcolor[HTML]{FADADD}
& \multicolumn{1}{l|}{S-FRET}   &
89.36 & 89.46 & 97.37 & 78.57 & 88.69 & 64.61 & 55.28 & 75.60 & 76.50 & 68.00 \\
\rowcolor[HTML]{FADADD}
& \multicolumn{1}{l|}{G-FRET}   &\textbf{89.75} & \textbf{89.97} & \textbf{97.49} & \textbf{85.52} & \textbf{90.68} & \textbf{66.17} & \textbf{57.87} & \textbf{76.93} & \underline{76.80} & \textbf{69.44} \\ 
\bottomrule
\end{tabular}
 \caption{At random seed 1, the accuracy comparison of different TTA methods on PACS and OfficeHome datasets based on ResNet-18 and ResNet-50 backbones. The best results are highlighted in \textbf{boldface}, and the second ones are \underline{underlined}.}
\label{tab:random1}
\end{center}
\end{table*}

\begin{table*}[h]
\begin{center}
\small
\begin{tabular}{@{}l|l|cccc>{\columncolor{blue!10}}c|cccc>{\columncolor{blue!10}}l@{}}
\toprule
\multirow{2}{*}{Backbone} & \multirow{2}{*}{Method}   & \multicolumn{4}{c}{PACS} & \multicolumn{1}{c}{\multirow{2}{*}{Avg}} & \multicolumn{4}{c}{OfficeHome} & \multicolumn{1}{c}{\multirow{2}{*}{Avg}} \\ \cmidrule(lr){3-6} \cmidrule(lr){8-11}
                          &                           & A              & C              & P              & S              & \multicolumn{1}{c}{}           & A              & C              & P              & R              & \multicolumn{1}{c}{}                     \\ \midrule
\multirow{9}{*}{ResNet-18} & \multicolumn{1}{l|}{Source \cite{he2016deep}
} &78.37 & 77.39 & 95.03 & 76.58 & 81.84 & 56.45 & 48.02 & 71.34 & 72.23 & 62.01 \\ 
& \multicolumn{1}{l|}{BN \cite{BN}
}     &81.05 & 80.42 & 94.97 & 73.73 & 82.54 & 55.71 & 49.10 & 70.85 & 72.55 & 62.05 \\ 
& \multicolumn{1}{l|}{SAR \cite{SAR}
}   &83.69 & 82.08 & 94.97 & 81.01 & 85.44 & 57.11 & 50.70 & 71.03 & 72.76 & 62.90 \\ 
& \multicolumn{1}{l|}{EATA \cite{EATA}
}   &80.91 & 81.06 & 94.97 & 72.77 & 82.43 & 56.49 & 49.10 & 71.95 & 73.22 & 62.69 \\ 
& \multicolumn{1}{l|}{TENT \cite{TENT}
}    &81.20 & 83.70 & 95.51 & 81.42 & 85.46 & 57.19 & 50.42 & 71.80 & 73.08 & 63.12 \\ 
& \multicolumn{1}{l|}{TSD \cite{TSD}
}   &87.06 & \underline{87.12} & 95.93 & 83.07 & \underline{88.29} & \textbf{57.48} & \underline{51.11} & 71.23 & 70.32 & 62.54 \\ 
& \multicolumn{1}{l|}{TEA \cite{TEA}
}    &\underline{87.09} & \textbf{87.97} & \textbf{96.47} & 81.50 & 88.26 & 56.53 & 50.42 & 71.91 & 72.96 & 62.96 \\ 
& \multicolumn{1}{l|}{TIPI \cite{TIPI}
}    &82.86 & 84.04 & 95.93 & \textbf{84.58} & 86.85 & 57.19 & 50.49 & \underline{72.00} & \underline{73.45} & \underline{63.28} \\ 
\rowcolor[HTML]{FADADD}
& \multicolumn{1}{l|}{S-FRET}   &
86.28 & 86.35 & 95.69 & 78.34 & 86.66 & 56.04 & 49.46 & 71.64 & 72.89 & 62.51 \\
\rowcolor[HTML]{FADADD}
& \multicolumn{1}{l|}{G-FRET}   &\textbf{87.45} & 87.07 & \underline{96.35} & \underline{83.66} & \textbf{88.63} & \underline{57.23} & \textbf{51.59} & \textbf{72.85} & \textbf{73.84} & \textbf{63.88} \\ 

 \midrule 
\multirow{9}{*}{ResNet-50} & \multicolumn{1}{l|}{Source \cite{he2016deep}
}&83.89 & 81.02 & 96.17 & 78.04 & 84.78 & 64.85 & 52.26 & 75.04 & 75.88 & 67.01 \\ 
& \multicolumn{1}{l|}{BN \cite{BN}
}     &84.72 & 85.20 & 96.59 & 72.72 & 84.80 & 63.33 & 53.08 & 73.64 & 74.82 & 66.22 \\ 
& \multicolumn{1}{l|}{SAR \cite{SAR}
}   &84.72 & 85.20 & 96.59 & 75.18 & 85.42 & 64.73 & 56.93 & 74.86 & 76.15 & 68.17 \\ 
& \multicolumn{1}{l|}{EATA \cite{EATA}
}   &85.45 & 85.11 & \textbf{96.47} & 72.38 & 84.85 & 63.86 & 54.73 & 74.00 & 75.60 & 67.05 \\ 
& \multicolumn{1}{l|}{TENT \cite{TENT}
}    &87.06 & 85.92 & 96.65 & 75.34 & 86.24 & 64.32 & 55.60 & 74.09 & 76.02 & 67.51 \\ 
& \multicolumn{1}{l|}{TSD \cite{TSD}
}   &\textbf{91.85} & \textbf{89.76} & \underline{97.49} & \underline{79.92} & \textbf{89.75} & 64.89 & \underline{57.46} & 75.76 & \underline{76.54} & 68.66 \\ 
& \multicolumn{1}{l|}{TEA \cite{TEA}
}    &88.57 & 88.27 & 97.07 & \textbf{80.48} & 88.60 & \underline{65.93} & 57.00 & \underline{75.96} & 76.25 & \underline{68.78} \\ 
& \multicolumn{1}{l|}{TIPI \cite{TIPI}
}    &\textbf{87.45} & 85.41 & 96.89 & 77.53 & 86.82 & 64.85 & 56.98 & 75.02 & 76.41 & 68.31 \\ 
\rowcolor[HTML]{FADADD}
& \multicolumn{1}{l|}{S-FRET}   &
90.77 & 88.69 & 97.31 & 77.07 & 88.46 & 64.44 & 55.40 & 75.26 & 76.29 & 67.85\\
\rowcolor[HTML]{FADADD}
& \multicolumn{1}{l|}{G-FRET}   &\textbf{91.85} & \underline{89.59} & \textbf{97.54} & 78.42 & \underline{89.35} & \textbf{66.05} & \textbf{57.73} & \textbf{76.17} & \textbf{77.25} & \textbf{69.30} \\ 
\bottomrule
\end{tabular}
 \caption{At random seed 2, the accuracy comparison of different TTA methods on PACS and OfficeHome datasets based on ResNet-18 and ResNet-50 backbones. The best results are highlighted in \textbf{boldface}, and the second ones are \underline{underlined}.}
\label{tab:random2}
\end{center}
\end{table*}

\begin{table*}[h]
\begin{center}
\small
\begin{tabular}{@{}l|l|cccc>{\columncolor{blue!10}}c|cccc>{\columncolor{blue!10}}l@{}}
\toprule
\multirow{2}{*}{Backbone} & \multirow{2}{*}{Method}   & \multicolumn{4}{c}{PACS} & \multicolumn{1}{c}{\multirow{2}{*}{Avg}} & \multicolumn{4}{c}{OfficeHome} & \multicolumn{1}{c}{\multirow{2}{*}{Avg}} \\ \cmidrule(lr){3-6} \cmidrule(lr){8-11}
                          &                           & A              & C              & P              & S              & \multicolumn{1}{c}{}           & A              & C              & P              & R              & \multicolumn{1}{c}{}                     \\ \midrule
\multirow{9}{*}{ResNet-18} & \multicolumn{1}{l|}{Source \cite{he2016deep}
} &78.37 & 77.39 & 95.03 & 76.58 & 81.84 & 56.45 & 48.02 & 71.34 & 72.23 & 62.01 \\ 
& \multicolumn{1}{l|}{BN \cite{BN}
}     &80.96 & 80.46 & 95.03 & 74.12 & 82.64 & 55.75 & 49.67 & 70.76 & 72.64 & 62.20 \\ 
& \multicolumn{1}{l|}{SAR \cite{SAR}
}   &83.40 & 81.02 & 95.03 & 81.01 & 85.11 & 57.27 & 50.68 & 70.80 & 73.08 & 62.96 \\ 
& \multicolumn{1}{l|}{EATA \cite{EATA}
}   &82.23 & 80.84 & 94.97 & 72.31 & 82.59 & 56.32 & 49.97 & 71.53 & 72.57 & 62.60 \\ 
& \multicolumn{1}{l|}{TENT \cite{TENT}
}    &83.15 & 79.91 & 95.27 & 82.08 & 85.10 & 57.03 & \underline{50.77} & 72.09 & \underline{73.40} & \underline{63.32} \\ 
& \multicolumn{1}{l|}{TSD \cite{TSD}
}   &86.77 & \underline{86.52} & 95.93 & 83.61 & 88.21 & \textbf{57.85} & 49.85 & 71.37 & 71.52 & 62.65 \\ 
& \multicolumn{1}{l|}{TEA \cite{TEA}
}    &\textbf{87.26} & 86.26 & \textbf{96.59} & 83.61 & \underline{88.43} & 56.70 & 50.42 & 71.75 & 72.69 & 62.89 \\ 
& \multicolumn{1}{l|}{TIPI \cite{TIPI}
}    &84.57 & 83.40 & 95.57 & \underline{83.69} & 86.81 & 56.94 & 50.72 & \underline{72.11} & 73.26 & 63.26 \\ 
\rowcolor[HTML]{FADADD}
& \multicolumn{1}{l|}{S-FRET}   &
86.18 & 86.35 & \underline{96.05} & 75.82 & 86.10 & 56.12 & 50.36 & 71.68 & 72.80 & 62.74\\
\rowcolor[HTML]{FADADD}
& \multicolumn{1}{l|}{G-FRET}   &\underline{86.87} & \textbf{87.33} & 95.87 & \textbf{84.07} & \textbf{88.53} & \underline{57.31} & \textbf{51.64} & \textbf{72.76} & \textbf{73.65} & \textbf{63.84} \\ 

 \midrule 
\multirow{9}{*}{ResNet-50} & \multicolumn{1}{l|}{Source \cite{he2016deep}
}&83.89 & 81.02 & 96.17 & 78.04 & 84.78 & 64.85 & 52.26 & 75.04 & 75.88 & 67.01 \\ 
& \multicolumn{1}{l|}{BN \cite{BN}
}     &85.16 & 85.20 & 96.71 & 72.18 & 84.81 & 63.41 & 52.88 & 73.26 & 75.58 & 66.28 \\ 
& \multicolumn{1}{l|}{SAR \cite{SAR}
}   &85.40 & 85.20 & 96.71 & 76.86 & 86.04 & 64.73 & 56.77 & 74.75 & 76.50 & 68.19 \\ 
& \multicolumn{1}{l|}{EATA \cite{EATA}
}   &85.21 & 85.49 & 96.41 & 72.10 & 84.80 & 64.15 & 53.36 & 74.41 & 75.42 & 66.83 \\ 
& \multicolumn{1}{l|}{TENT \cite{TENT}
}    &86.18 & 86.90 & 96.95 & 79.97 & 87.50 & 64.32 & 55.51 & 74.93 & 76.27 & 67.76 \\ 
& \multicolumn{1}{l|}{TSD \cite{TSD}
}   &\underline{90.48} & \textbf{90.32} & \underline{97.78} & 81.98 & \underline{90.14} & \underline{65.31} & 56.59 & 75.74 & 76.75 & 68.60 \\ 
& \multicolumn{1}{l|}{TEA \cite{TEA}
}    &88.13 & 86.95 & 97.07 & \underline{82.41} & 88.64 & 65.27 & \textbf{58.56} & \underline{76.23} & \textbf{77.39} & \textbf{69.36} \\ 
& \multicolumn{1}{l|}{TIPI \cite{TIPI}
}    &86.77 & 87.54 & 97.07 & 81.04 & 88.10 & 64.65 & 56.72 & 75.20 & 77.09 & 68.42 \\ 
\rowcolor[HTML]{FADADD}
& \multicolumn{1}{l|}{S-FRET}   &
89.31 & 88.95 & 97.72 & 78.16 & 88.54 & 64.61 & 55.10 & 75.38 & 76.59 & 67.92\\
\rowcolor[HTML]{FADADD}
& \multicolumn{1}{l|}{G-FRET}   &\textbf{90.82} & \underline{90.23} & \textbf{97.90} & \textbf{83.46} & \textbf{90.60} & \textbf{65.47} & \underline{57.34} & \textbf{76.28} & \underline{77.32} & \underline{69.10} \\ 
\bottomrule
\end{tabular}
 \caption{At random seed 3, the accuracy comparison of different TTA methods on PACS and OfficeHome datasets based on ResNet-18 and ResNet-50 backbones. The best results are highlighted in \textbf{boldface}, and the second ones are \underline{underlined}.}
\label{tab:random3}
\end{center}
\end{table*}

\begin{table*}[h]
\begin{center}
\small
\begin{tabular}{@{}l|l|cccc>{\columncolor{blue!10}}c|cccc>{\columncolor{blue!10}}l@{}}
\toprule
\multirow{2}{*}{Backbone} & \multirow{2}{*}{Method}   & \multicolumn{4}{c}{PACS} & \multicolumn{1}{c}{\multirow{2}{*}{Avg}} & \multicolumn{4}{c}{OfficeHome} & \multicolumn{1}{c}{\multirow{2}{*}{Avg}} \\ \cmidrule(lr){3-6} \cmidrule(lr){8-11}
                          &                           & A              & C              & P              & S              & \multicolumn{1}{c}{}           & A              & C              & P              & R              & \multicolumn{1}{c}{}                     \\ \midrule
\multirow{9}{*}{ResNet-18} & \multicolumn{1}{l|}{Source \cite{he2016deep}
} &78.37 & 77.39 & 95.03 & 76.58 & 81.84 & 56.45 & 48.02 & 71.34 & 72.23 & 62.01 \\ 
& \multicolumn{1}{l|}{BN \cite{BN}
}     &81.10 & 81.06 & 95.27 & 73.63 & 82.77 & 55.38 & 49.28 & 70.65 & 72.39 & 61.92 \\ 
& \multicolumn{1}{l|}{SAR \cite{SAR}
}   &83.40 & 82.04 & 95.27 & 80.91 & 85.40 & \underline{58.06} & \underline{50.79} & 71.39 & 72.66 & 63.23 \\ 
& \multicolumn{1}{l|}{EATA \cite{EATA}
}   &81.49 & 79.65 & 95.51 & 74.68 & 82.83 & 55.87 & 49.92 & 71.34 & 73.22 & 62.59 \\ 
& \multicolumn{1}{l|}{TENT \cite{TENT}
}    &85.69 & 83.83 & \underline{96.05} & 80.73 & 86.58 & 57.31 & 50.17 & \underline{72.61} & \textbf{73.42} & \underline{63.38} \\ 
& \multicolumn{1}{l|}{TSD \cite{TSD}
}   &\textbf{87.84} & \textbf{87.93} & 95.75 & 83.41 & \underline{88.73} & \textbf{58.10} & 50.22 & 71.10 & 70.87 & 62.57 \\ 
& \multicolumn{1}{l|}{TEA \cite{TEA}
}    &86.57 & 86.05 & \textbf{96.13} & 82.72 & 87.87 & 56.70 & 50.22 & 71.57 & 72.85 & 62.83 \\ 
& \multicolumn{1}{l|}{TIPI \cite{TIPI}
}    &85.89 & 85.62 & 95.87 & \underline{83.84} & 87.80 & 57.19 & 50.10 & 72.40 & \textbf{73.42} & 63.28 \\ 
\rowcolor[HTML]{FADADD}
& \multicolumn{1}{l|}{S-FRET}   &
86.67 & 87.24 & 95.81 & 78.29 & 87.00 & 56.41 & 49.90 & 71.66 & 72.71 & 62.67 \\
\rowcolor[HTML]{FADADD}
& \multicolumn{1}{l|}{G-FRET}   &\underline{87.60} & \underline{87.54} & 95.93 & \textbf{85.06} & \textbf{89.03} & 57.97 & \textbf{51.20} & \textbf{72.99} & 73.38 & \textbf{63.89} \\ 

 \midrule 
\multirow{9}{*}{ResNet-50} & \multicolumn{1}{l|}{Source \cite{he2016deep}
}&83.89 & 81.02 & 96.17 & 78.04 & 84.78 & 64.85 & 52.26 & 75.04 & 75.88 & 67.01 \\ 
& \multicolumn{1}{l|}{BN \cite{BN}
}     &85.55 & 85.58 & 96.29 & 72.41 & 84.96 & 63.21 & 52.90 & 73.55 & 75.07 & 66.18 \\ 
& \multicolumn{1}{l|}{SAR \cite{SAR}
}   &85.55 & 85.58 & 96.29 & 76.89 & 86.08 & 64.61 & 55.56 & 74.79 & 75.97 & 67.73 \\ 
& \multicolumn{1}{l|}{EATA \cite{EATA}
}   &85.01 & 85.58 & 96.77 & 72.26 & 84.90 & 64.52 & 54.46 & 74.30 & 76.47 & 67.44 \\ 
& \multicolumn{1}{l|}{TENT \cite{TENT}
}    &87.79 & 86.69 & 96.83 & 79.41 & 87.68 & 64.61 & 54.27 & 74.59 & 76.11 & 67.39 \\ 
& \multicolumn{1}{l|}{TSD \cite{TSD}
}   &\underline{91.65} & \textbf{90.10} & \underline{97.19} & 80.33 & \underline{89.82} & \underline{65.47} & 57.00 & \underline{76.53} & 76.73 & \underline{68.93} \\ 
& \multicolumn{1}{l|}{TEA \cite{TEA}
}    &88.57 & 87.59 & \textbf{97.37} & \underline{80.40} & 88.48 & 65.43 & \underline{57.02} & 75.92 & 76.54 & 68.73 \\ 
& \multicolumn{1}{l|}{TIPI \cite{TIPI}
}    &88.53 & 86.99 & 96.89 & 79.94 & 88.09 & 64.98 & 56.29 & 75.20 & \underline{76.91} & 68.34 \\ 
\rowcolor[HTML]{FADADD}
& \multicolumn{1}{l|}{S-FRET}   &
90.09 & 88.40 & 97.07 & 77.14 & 88.17 & 64.28 & 55.03 & 75.83 & 76.45 & 67.90 \\
\rowcolor[HTML]{FADADD}
& \multicolumn{1}{l|}{G-FRET}   &\textbf{91.80} & \underline{90.06} & \underline{97.19} & \textbf{81.93} & \textbf{90.24} & \textbf{66.50} & \textbf{58.08} & \textbf{76.71} & \textbf{77.23} & \textbf{69.63} \\ 
\bottomrule
\end{tabular}
 \caption{At random seed 4, the accuracy comparison of different TTA methods on PACS and OfficeHome datasets based on ResNet-18 and ResNet-50 backbones. The best results are highlighted in \textbf{boldface}, and the second ones are \underline{underlined}.}
\label{tab:random4}
\end{center}
\end{table*}

\begin{table*}[h]
\begin{center}
\small
\resizebox{\textwidth}{!}{
\begin{tabular}{@{}ll|ccccccccccccccc>{\columncolor{blue!10}}l@{}}
\toprule
& \multirow{2}{*}{Method}   & \multicolumn{15}{c}{$t  \xrightarrow{\hspace{16cm}}$}&\multicolumn{1}{c}{\multirow{2}{*}{Avg}} \\ \cmidrule(lr){3-17}&                           & Gau.& Sho.& Imp.& Def.& \multicolumn{1}{c}{Gla.}           & Mot.& Zoo.& Sno.& Fro.&  Fog&   Bri.&Con.&Ela.&  Pix.&Jpe.&\multicolumn{1}{c}{} \\ \midrule
& \multicolumn{1}{l|}{Source \cite{he2016deep}} &$27.43$& $33.56$& $21.57$& $43.64$& $40.48$& $51.26$& $51.29$& $68.18$& $54.52$&  $66.65$&   $87.50$&$27.59$&$67.06$&  $48.86$&$72.37$&$50.80$\\
                          & \multicolumn{1}{l|}{BN \cite{BN}}     & $66.30$& $68.18$& $57.13$& $82.50$& $57.44$& $79.73$& $81.98$& $74.83$& $74.12$&  $78.91$&   $86.96$&$82.02$&$70.23$&  $73.43$&$70.94$&$73.65$\\
                          & \multicolumn{1}{l|}{TENT \cite{TENT}}    & $67.26$& $71.46$& $61.21$& $84.07$& $61.37$& $\underline{81.66}$& $84.36$& $78.18$& $77.55$&  $\underline{80.14}$&   $88.44$&$81.41$&$73.54$&  $78.53$&$76.19$&$76.36$\\
                          & \multicolumn{1}{l|}{EATA \cite{EATA}}   & $66.39$& $68.50$& $57.32$& $82.52$& $57.42$& $79.94$& $82.09$& $74.80$& $74.14$&  $78.90$&   $86.98$&$81.93$&$70.10$&  $73.62$&$70.88$&$73.70$\\
                          & \multicolumn{1}{l|}{SAR \cite{SAR}}   & $66.46$& $68.24$& $57.47$& $82.52$& $57.83$& $79.76$& $81.98$& $74.83$& $74.29$&  $78.92$&   $86.96$&$\underline{82.36}$&$70.26$&  $73.43$&$70.94$&$73.75$\\
                          & \multicolumn{1}{l|}{TIPI \cite{TIPI}}    & $\underline{67.57}$& $\textbf{72.14}$& $\textbf{62.88}$& $\underline{84.19}$& $\textbf{63.55}$& $81.63$& $\underline{84.44}$& $\textbf{79.06}$& $\textbf{79.07}$&  $79.61$&   $\underline{88.68}$&$81.92$&$\textbf{75.33}$&  $\textbf{79.92}$&$\textbf{78.11}$&$\underline{77.21}$\\
                          & \multicolumn{1}{l|}{TEA \cite{TEA}}    & $66.71$& $69.24$& $59.46$& $82.78$& $59.98$& $80.87$& $82.88$& $76.40$& $75.60$&  $79.82$&   $86.72$&$81.48$&$71.89$&  $74.91$&$72.91$&$74.78 $\\
                          & \multicolumn{1}{l|}{TSD \cite{TSD}}   & $66.97$& $70.31$& $60.63$& $83.24$& $61.10$& $81.52$& $83.97$& $77.15$& $76.75$&  $80.08$&   $86.76$&$80.42$&$72.66$&  $76.43$&$73.42$&$75.43$\\
                          \rowcolor[HTML]{FADADD}
                          & \multicolumn{1}{l|}{S-FRET}   &
                          $66.52$ & $69.45$ & $59.40$ & $82.78$ & $59.43$ & $80.64$ & $82.98$ & $76.05$ & $75.73$ & $79.59$ & $86.92$ & $80.68$ & $71.56$ & $75.11$ & $72.61$ & $74.63$ \\
                          \rowcolor[HTML]{FADADD}
                          & \multicolumn{1}{l|}{G-FRET}   & $\textbf{67.79}$& $\underline{71.83}$& $\underline{62.81}$& $\textbf{84.31}$& $\underline{62.63}$& $\textbf{82.07}$& $\textbf{84.89}$& $\underline{78.91}$& $\underline{79.00}$&  $\textbf{81.01}$&   $\textbf{88.87}$&$\textbf{82.69}$&$\underline{74.86}$&  $\underline{79.47}$&$\underline{77.74}$&$\textbf{77.26}$\\
\end{tabular}
}
\caption{Accuracy comparisons of different TTA methods on CIFAR-10-C dataset at damage level of 5, with 15 types of damage applied sequentially to a continuously adapted model. The best results are highlighted in \textbf{boldface}, and the second ones are \underline{underlined}.}
\label{tab:CIFAR10}
\end{center}
\end{table*}

\begin{table*}[h]
\begin{center}
\small
\resizebox{\textwidth}{!}{
\begin{tabular}{@{}ll|ccccccccccccccc>{\columncolor{blue!10}}l@{}}
\toprule
& \multirow{2}{*}{Method}   & \multicolumn{15}{c}{$t  \xrightarrow{\hspace{16cm}}$}&\multicolumn{1}{c}{\multirow{2}{*}{Avg}} \\ \cmidrule(lr){3-17}&                           & Gau.& Sho.& Imp.& Def.& \multicolumn{1}{c}{Gla.}           & Mot.& Zoo.& Sno.& Fro.&  Fog&   Bri.&Con.&Ela.&  Pix.&Jpe.&\multicolumn{1}{c}{} \\ \midrule
& \multicolumn{1}{l|}{Source \cite{he2016deep}} &$10.46$ & $12.49$ & $3.36$ & $34.44$ & $23.63$ & $38.10$ & $42.67$ & $39.25$ & $33.01$ & $32.84$ & $55.78$ & $11.55$ & $46.48$ & $34.88$ & $46.15$ & $31.01$\\
                          & \multicolumn{1}{l|}{BN \cite{BN}}    &$39.84$ & $39.51$ & $29.88$ & $56.43$ & $41.08$ & $54.34$ & $58.82$ & $48.52$ & $49.36$ & $46.60$ & $61.82$ & $48.78$ & $49.92$ & $54.17$ & $45.39$ & $48.30$ \\
                          & \multicolumn{1}{l|}{TENT \cite{TENT}}    &$40.61$ & $41.94$ & $32.09$ & $\underline{57.84}$ & $\underline{44.35}$ & $56.57$ & $\underline{60.96}$ & $\underline{51.51}$ & $52.03$ & $\underline{49.23}$ & $\textbf{63.94}$ & $49.62$ & $53.53$ & $\textbf{57.60}$ & $50.03$ & $\underline{50.79}$\\
                          & \multicolumn{1}{l|}{EATA \cite{EATA}}   &$40.59$ & $41.82$ & $32.58$ & $\textbf{57.97}$ & $43.43$ & $\underline{56.76}$ & $60.33$ & $50.79$ & $51.90$ & $48.71$ & $\underline{63.83}$ & $\underline{49.88}$ & $\underline{53.96}$ & $\textbf{57.60}$ & $50.50$ & $50.71$ \\
                          & \multicolumn{1}{l|}{SAR \cite{SAR}}   &$40.09$ & $40.67$ & $31.52$ & $57.01$ & $42.16$ & $55.63$ & $59.54$ & $50.53$ & $50.31$ & $47.73$ & $62.50$ & $43.27$ & $51.03$ & $55.49$ & $48.80$ & $49.09$\\
                          & \multicolumn{1}{l|}{TIPI \cite{TIPI}}    &$\underline{40.62}$ & $\underline{42.29}$ & $\underline{32.67}$ & $57.06$ & $\textbf{44.84}$ & $55.45$ & $59.58$ & $\textbf{52.19}$ & $\underline{52.15}$ & $46.33$ & $61.91$ & $43.60$ & $52.20$ & $57.39$ & $\underline{50.67}$ & $49.93$\\
                          & \multicolumn{1}{l|}{TEA \cite{TEA}}    &$40.13$ & $39.90$ & $30.82$ & $56.28$ & $41.48$ & $54.73$ & $59.16$ & $48.79$ & $49.31$ & $46.26$ & $61.41$ & $48.48$ & $50.22$ & $54.03$ & $46.63$ & $48.51$\\
                          & \multicolumn{1}{l|}{TSD \cite{TSD}}   &$39.84$ & $39.65$ & $30.14$ & $56.63$ & $41.17$ & $54.65$ & $59.03$ & $48.71$ & $49.71$ & $47.25$ & $61.95$ & $48.84$ & $50.65$ & $54.78$ & $46.36$ & $48.62$ \\
                          \rowcolor[HTML]{FADADD}
                          & \multicolumn{1}{l|}{S-FRET}   &
                          $39.84$ & $39.94$ & $31.10$ & $56.66$ & $42.01$ & $55.19$ & $59.46$ & $48.66$ & $49.21$ & $47.21$ & $61.52$ & $46.80$ & $50.66$ & $53.88$ & $45.75$ & $48.53$ \\
                          \rowcolor[HTML]{FADADD}
                          & \multicolumn{1}{l|}{G-FRET}   &$\textbf{41.08}$ & $\textbf{43.23}$ & $\textbf{33.69}$ & $57.49$ & $44.07$ & $\textbf{56.87}$ & $\textbf{61.16}$ & $51.24$ & $\textbf{52.18}$ & $\textbf{49.25}$ & $63.12$ & $\textbf{50.04}$ & $\textbf{54.09}$ & $57.42$ & $\textbf{51.02}$ & $\textbf{51.06}$\\
\end{tabular}
}
\caption{Accuracy comparisons of different TTA methods on CIFAR-100-C dataset at damage level of 5, with 15 types of damage applied sequentially to a continuously adapted model. The best results are highlighted in \textbf{boldface}, and the second ones are \underline{underlined}.}
\label{tab:CIFAR100}
\end{center}
\end{table*}

\begin{table*}[h]
\begin{center}
\small
\resizebox{\textwidth}{!}{
\begin{tabular}{@{}ll|ccccccccccccccc>{\columncolor{blue!10}}l@{}}
\toprule
& \multirow{2}{*}{Method}   & \multicolumn{15}{c}{ImageNet-C}&\multicolumn{1}{c}{\multirow{2}{*}{Avg}} \\ \cmidrule(lr){3-17}&                           & Gau.& Sho.& Imp.& Def.& \multicolumn{1}{c}{Gla.}           & Mot.& Zoo.& Sno.& Fro.&  Fog&   Bri.&Con.&Ela.&  Pix.&Jpe.&\multicolumn{1}{c}{} \\ \midrule
& \multicolumn{1}{l|}{Source \cite{he2016deep}} &$1.54$ & $2.27$ & $1.48$ & $11.44$ & $8.68$ & $11.12$ & $17.62$ & $10.64$ & $16.21$ & $14.02$ & $51.52$ & $3.44$ & $16.49$ & $23.35$ & $30.67$ & $14.70$ \\
                          & \multicolumn{1}{l|}{BN \cite{BN}}    &$13.65$ & $14.84$ & $14.17$ & $11.95$ & $13.04$ & $23.34$ & $33.89$ & $29.18$ & $28.42$ & $40.80$ & $58.11$ & $12.09$ & $38.92$ & $44.35$ & $37.08$ & $27.59$ \\
                          & \multicolumn{1}{l|}{TENT \cite{TENT}}    &$23.45$ & $25.71$ & $24.08$ & $18.79$ & $20.90$ & $33.54$ & $42.85$ & $39.64$ & $32.95$ & $50.36$ & $60.13$ & $10.68$ & $48.81$ & $51.96$ & $46.98$ & $35.39$ \\
                          & \multicolumn{1}{l|}{EATA \cite{EATA}}   &$\textbf{28.24}$ & $\textbf{30.16}$ & $\textbf{28.88}$ & $\textbf{25.30}$ & $\textbf{25.74}$ & $\textbf{36.61}$ & $\underline{43.71}$ & $\textbf{41.80}$ & $36.42$ & $\underline{50.87}$ & $59.12$ & $\textbf{31.75}$ & $\underline{49.10}$ & $\underline{52.33}$ & $\underline{47.82}$ & $\textbf{39.19}$ \\
                          & \multicolumn{1}{l|}{SAR \cite{SAR}}   &$\underline{28.04}$ & $\underline{29.59}$ & $\underline{27.88}$ & $\underline{23.66}$ & $\underline{23.90}$ & $\underline{36.16}$ & $43.40$ & $\underline{40.94}$ & $\textbf{36.71}$ & $\textbf{51.01}$ & $\underline{60.18}$ & $\underline{27.38}$ & $48.95$ & $\textbf{52.47}$ & $\textbf{47.98}$ & $\underline{38.55}$ \\
                          & \multicolumn{1}{l|}{TIPI \cite{TIPI}}    &$24.45$ & $26.52$ & $24.75$ & $20.37$ & $22.25$ & $33.65$ & $42.46$ & $39.31$ & $33.47$ & $49.93$ & $59.44$ & $12.53$ & $48.41$ & $51.51$ & $46.92$ & $35.73$ \\
                          & \multicolumn{1}{l|}{TEA \cite{TEA}}    &$18.82$ & $20.50$ & $19.00$ & $16.27$ & $17.68$ & $28.51$ & $39.17$ & $35.19$ & $32.26$ & $46.92$ & $59.16$ & $15.42$ & $44.39$ & $48.81$ & $43.64$ & $32.38$ \\
                          & \multicolumn{1}{l|}{TSD \cite{TSD}}   &$15.60$ & $16.99$ & $16.13$ & $15.59$ & $15.41$ & $28.69$ & $38.07$ & $32.92$ & $30.01$ & $45.90$ & $58.69$ & $7.62$ & $41.06$ & $47.47$ & $41.52$ & $30.11$ \\
                          \rowcolor[HTML]{FADADD}
                          & \multicolumn{1}{l|}{S-FRET}   &
                          $15.04$ & $16.40$ & $15.57$ & $13.77$ & $14.67$ & $25.65$ & $36.00$ & $31.08$ & $29.08$ & $43.13$ & $58.64$ & $12.33$ & $40.37$ & $45.76$ & $39.07$ & $29.10$ \\
                          \rowcolor[HTML]{FADADD}
                          & \multicolumn{1}{l|}{G-FRET}   &$24.85$ & $27.47$ & $25.49$ & $20.82$ & $22.71$ & $35.10$ & $\textbf{43.76}$ & $40.66$ & $\underline{36.68}$ & $50.80$ & $\textbf{60.3}$ & $14.20$ & $\textbf{49.28}$ & $52.24$ & $47.52$ & $36.79$ \\
\end{tabular}
}
\caption{Accuracy comparisons of different TTA methods on ImageNet-C dataset at damage level of 5, with 15 types of damage applied independently to the adapted model based on ResNet-18. The best results are highlighted in \textbf{boldface}, and the second ones are \underline{underlined}.}
\label{tab:ImageNet}
\end{center}
\end{table*}

\begin{table*}[t]
\begin{center}
\small
\resizebox{0.6\textwidth}{!}{
    \begin{tabular}{lccccc}
    \toprule
    \multirow{2}{*}{Method}& \multicolumn{4}{c}{VLCS}&\multirow{2}{*}{Avg}\\
 \cmidrule(lr){2-5} & C& L& S& V&\\ \midrule
    ResNet-18& 94.49& 60.96& 67.73& 71.50&73.67\\
    \rowcolor[HTML]{FADADD}
    +S-FRET & 96.11& 59.71& 67.28& 72.51&73.90 \\
    \rowcolor[HTML]{FADADD}
    +G-FRET & \textbf{96.68}& \textbf{64.42}& \textbf{67.82}& \textbf{73.99}&\textbf{75.73}\\ \midrule
    ResNet-50& 95.55& \textbf{60.77}& 71.12& 72.16&74.90\\
    \rowcolor[HTML]{FADADD}
    +S-FRET & 96.05& 57.89& 69.86& \textbf{78.10}&75.48 \\
    \rowcolor[HTML]{FADADD}
    +G-FRET & \textbf{96.96}& 58.51& \textbf{72.27}& 77.93&\textbf{76.42}\\ \midrule
    ViT-B/16& 97.81& 64.38& 69.71& 73.84&76.44\\
    \rowcolor[HTML]{FADADD}
    +S-FRET & 97.81& 67.32& 69.59& 73.99&77.18 \\
    \rowcolor[HTML]{FADADD}
    +G-FRET & \textbf{98.52}& \textbf{68.00}& \textbf{73.49}& \textbf{74.23}&\textbf{78.56}\\ \midrule
    \end{tabular}
    } 
\caption{Accuracy on the VLCS dataset with different backbones: ResNet-18/50 and ViT-B/16. }
\label{tab:VLCS}
\end{center}
\end{table*}

\begin{table*}[t]
\begin{center}
\small
\resizebox{0.8\textwidth}{!}{
    \begin{tabular}{lccccccc}
    \toprule
    \multirow{2}{*}{Method}& \multicolumn{6}{c}{DomainNet}&\multirow{2}{*}{Avg}\\
 \cmidrule(lr){2-7}& C& I& P& Q& R&S&\\ \midrule
    ResNet-18& 57.30& \textbf{16.86}& 45.03& 12.69& 56.89&46.00&39.13\\
    \rowcolor[HTML]{FADADD}
    +S-FRET & 57.69& 12.58& 44.55& 15.18& \textbf{57.71}&47.86&39.26\\
    \rowcolor[HTML]{FADADD}
    +G-FRET & \textbf{58.97}& 14.10& \textbf{46.16}& \textbf{15.22}& 57.42&\textbf{49.48}&\textbf{40.22}\\ \midrule
    ResNet-50& 63.68& \textbf{20.93}& 50.35& 12.95& 62.16&51.42&43.58\\
    \rowcolor[HTML]{FADADD}
    +S-FRET & 63.95& 15.72& 50.00& \textbf{15.23}& \textbf{63.51}&53.09&43.59\\
    \rowcolor[HTML]{FADADD}
    +G-FRET & \textbf{64.85}& 17.61& \textbf{51.19}& 14.71& 63.33&\textbf{53.61}&\textbf{44.22}\\ \midrule
    ViT-B/16& 71.91& 25.56& 55.95& 18.36& 70.66&57.45&49.98\\
    \rowcolor[HTML]{FADADD}
    +S-FRET & 72.30& \textbf{27.17}& \textbf{59.45}& 17.25& 71.48&59.65&51.22\\
    \rowcolor[HTML]{FADADD}
    +G-FRET & \textbf{72.63}& 26.04& 58.28& \textbf{18.92}& \textbf{72.61}&\textbf{60.50}&\textbf{51.50}\\ \midrule
    \end{tabular}
    } 
\caption{Accuracy on the DomainNet dataset with different backbones: ResNet-18/50 and ViT-B/16. }
\label{tab:DomainNet}
\end{center}
\end{table*}

\begin{table*}[t]
\begin{center}
\small
\resizebox{\textwidth}{!}{
    \begin{tabular}{lcccccccccccccccc}
    \toprule
    \multirow{2}{*}{Method}& \multicolumn{15}{c}{ImageNet-C}&\multirow{2}{*}{Avg}\\
 \cmidrule(lr){2-16}& Gau.& Sho.& Imp.& Def.& Gla.&Mot.& Zoo.& Sno.& Fro.& Fog& Bri.& Con.& Ela.& Pix.&Jpe.&\\ \midrule
    ResNet-18&$1.54$ & $2.27$ & $1.48$ & $11.44$ & $8.68$ & $11.12$ & $17.62$ & $10.64$ & $16.21$ & $14.02$ & $51.52$ & $3.44$ & $16.49$ & $23.35$ & $30.67$ & $14.70$ \\
    \rowcolor[HTML]{FADADD}
    +S-FRET &
    $15.04$ & $16.40$ & $15.57$ & $13.77$ & $14.67$ & $25.65$ & $36.00$ & $31.08$ & $29.08$ & $43.13$ & $58.64$ & $12.33$ & $40.37$ & $45.76$ & $39.07$ & $29.10$ \\
    \rowcolor[HTML]{FADADD}
    +G-FRET &$\textbf{24.85}$ & $\textbf{27.47}$ & $\textbf{25.49}$ & $\textbf{20.82}$ & $\textbf{22.71}$ & $\textbf{35.10}$ & $\textbf{43.76}$ & $\textbf{40.66}$ & $\textbf{36.68}$ & $\textbf{50.80}$ & $\textbf{60.30}$ & $\textbf{14.20}$ & $\textbf{49.28}$ & $\textbf{52.24}$ & $\textbf{47.52}$ & $\textbf{36.79}$ \\ \midrule
    ResNet-50&$3.00$ & $3.70$ & $2.64$ & $17.91$ & $9.74$ & $14.71$ & $22.45$ & $16.60$ & $23.06$ & $24.01$ & $59.12$ & $5.38$ & $16.51$ & $20.87$ & $32.63$ & $18.15$ \\
    \rowcolor[HTML]{FADADD}
    +S-FRET &
    $19.26$ & $18.81$ & $19.91$ & $18.66$ & $18.16$ & $29.96$ & $43.32$ & $38.29$ & $34.02$ & $51.74$ & $66.52$ & $16.69$ & $47.22$ & $52.51$ & $44.60$ & $34.65$ \\
    \rowcolor[HTML]{FADADD}
    +G-FRET &$\textbf{29.22}$ & $\textbf{29.13}$ & $\textbf{29.83}$ & $\textbf{25.99}$ & $\textbf{26.68}$ & $\textbf{44.10}$ & $\textbf{50.83}$ & $\textbf{49.15}$ & $\textbf{43.40}$ & $\textbf{58.43}$ & $\textbf{67.35}$ & $\textbf{17.88}$ & $\textbf{57.12}$ & $\textbf{59.68}$ & $\textbf{53.76}$ & $\textbf{42.84}$ \\ \midrule
    ViT-B/16&$35.09$ & $32.16$ & $35.88$ & $31.42$ & $25.31$ & $39.45$ & $31.55$ & $24.47$ & $30.13$ & $54.74$ & $64.48$ & $48.98$ & $34.20$ & $53.17$ & $56.45$ & $39.83$ \\
    \rowcolor[HTML]{FADADD}
    +S-FRET &
    $51.62$ & $53.02$ & $53.54$ & $49.54$ & $50.21$ & $56.68$ & $\textbf{59.19}$ & $\textbf{61.93}$ & $\textbf{61.22}$ & $70.65$ & $72.85$ & $65.43$ & $66.11$ & $67.87$ & $65.61$ & $60.36$ \\
    \rowcolor[HTML]{FADADD}
    +G-FRET &$\textbf{57.56}$ & $\textbf{56.80}$ & $\textbf{58.02}$ & $\textbf{56.86}$ & $\textbf{57.42}$ & $\textbf{62.11}$ & $58.98$ & $41.25$ & $59.39$ & $\textbf{72.69}$ & $\textbf{77.31}$ & $\textbf{70.33}$ & $\textbf{66.61}$ & $\textbf{71.93}$ & $\textbf{70.25}$ & $\textbf{62.50}$ \\ \midrule
    \end{tabular}
    } 
    \caption{Accuracy on the ImageNet-C dataset with different backbones: ResNet-18/50 and ViT-B/16. }
\label{tab:ImageNet2}
\end{center}
\end{table*}

\end{document}